\documentclass[11pt, a4paper, logo, copyright]{googledeepmind}

\usepackage[authoryear, sort&compress, round]{natbib}
\usepackage{microtype}
\usepackage{graphicx}
\usepackage{caption}
\usepackage{subcaption}
\usepackage{booktabs} 

\usepackage{hyperref}
\usepackage{multirow}
\usepackage{float}

\usepackage{amsmath}
\usepackage{amssymb}
\usepackage{mathtools}
\usepackage{amsthm}

\usepackage[capitalize,noabbrev]{cleveref}

\usepackage[textsize=tiny]{todonotes}
\usepackage{enumitem} 

\theoremstyle{plain}

\theoremstyle{definition}

\theoremstyle{remark}

\newcommand{\eat}[1]{}

\newcommand{\gsmname}{R-GSM}

\title{Premise Order Matters in Reasoning with Large Language Models}
\usepackage{inconsolata}

\author[1 *]{Xinyun Chen}
\author[1 2 *]{Ryan A. Chi}
\author[1]{Xuezhi Wang}
\author[1]{Denny Zhou}

\affil[*]{Equal contribution}
\affil[1]{Google DeepMind}
\affil[2]{Stanford University
\newline
\texttt{\{xinyunchen,xuezhiw,dennyzhou\}@google.com, ryanchi@cs.stanford.edu}
}

\begin{abstract}
Large language models (LLMs) have accomplished remarkable reasoning performance in various domains.
However, in the domain of reasoning tasks, we discover a frailty: LLMs are surprisingly brittle to the \textit{ordering of the premises}, despite the fact that such ordering does not alter the underlying task.
In particular, we observe that LLMs achieve the best performance when the premise order aligns with the context required in intermediate reasoning steps.
For example, in deductive reasoning tasks,  presenting the premises in the same order as the ground truth proof in the prompt (as opposed to random ordering) drastically increases the model's accuracy.
We first examine the effect of premise ordering on deductive reasoning on a variety of LLMs, and our evaluation shows that permuting the premise order can cause a performance drop of over 30\%.
In addition, we release the benchmark \textbf{\gsmname{}}, based on GSM8K, to examine the ordering effect for mathematical problem-solving, and we again observe a significant drop in accuracy, relative to the original GSM8K benchmark.
\end{abstract}

\begin{document}

\maketitle

\begin{figure*}[htbp!]
    \centering
    \includegraphics[width=\textwidth]{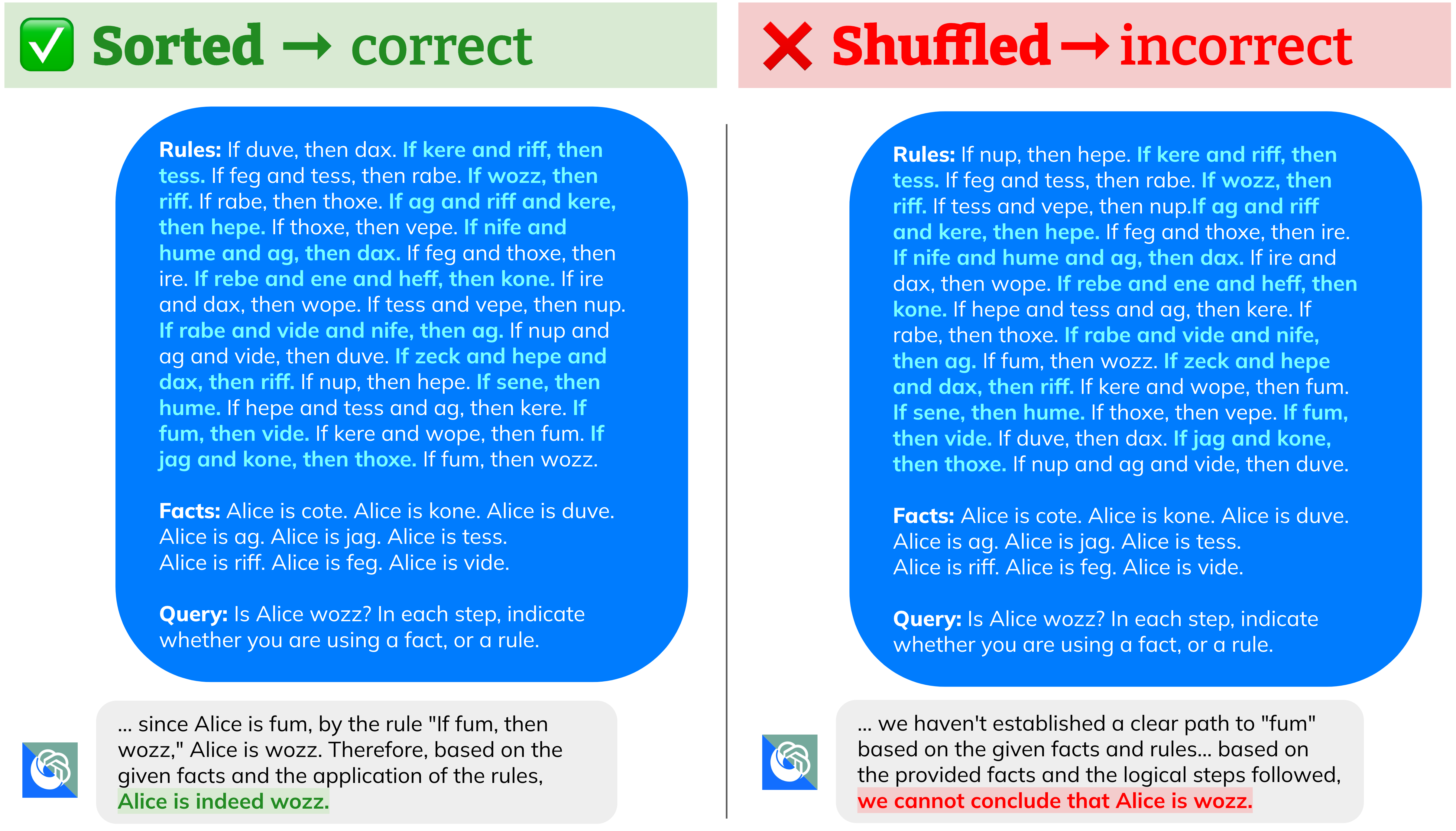}
    \caption{Premise order affects the reasoning performance: a failure case for logical reasoning. Left: rules are sorted in the same order as the ground truth proof (forward order with $\tau=1$ as defined in Section~\ref{sec:benchmark-logic}). Right: the wrong prediction with GPT-4-turbo after shuffling the rule set ($\tau=0$). Distracting rules are in bold and light blue. }
    \label{fig:headline} 
\end{figure*}

\section{Introduction}
\label{sec:intro}

Large language models (LLMs) have demonstrated impressive performance across a variety of reasoning tasks~\citep{wei2022chain,cobbe2021training,hendrycks2021measuring,chen2021evaluating,austin2021program}. In particular, recent state-of-the-art LLMs have reached or even surpassed human performance on multiple reasoning benchmarks, including STEM problem-solving and code generation~\citep{bubeck2023sparks,team2023gemini,li2022competition}. However, recent works show that LLMs exhibit failure modes that align with human-like cognitive bias~\citep{berglund2023reversal,shi2023large,hagendorff2023human,jones2022capturing,mccoy2023embers}. For example, \citet{berglund2023reversal} revealed the \textit{Reversal Curse}; i.e., LLMs trained on ``A is B'' tend to fail to infer that ``B is A.'' Distractibility is another failure mode \citep{shi2023large,jones2022capturing}, where the LLM performance drastically decreases when irrelevant context is included in the task description.

In this work, we investigate the effect that premise order has on LLM reasoning. Specifically, in deductive reasoning, changing the order of premises alone does not change the conclusion. Consider the following illustrative example:
\begin{enumerate}
    \item If $A$ then $B$.
    \item If $B$ then $C$.
    \item $A$ is \texttt{True}.
\end{enumerate} We can derive that $C$ is \texttt{True} regardless of the order of these 3 premises. While some studies show that humans have a preference on the premise order to facilitate their reasoning~\citep{dekeyser2000preferred,girotto1997effect}, the premise order does not drastically affect human performance, especially for problems that only involve \textit{modus ponens} (if P then Q; P; therefore Q), which are relatively straightforward for humans.

In contrast to humans, we observe that for LLMs, the premise order has a significant impact on reasoning performance. In particular, LLMs reach the best performance when the premises are arranged \textbf{in the same order} as they appear in the ground-truth proof. Taking the illustrative problem above as an example, we observe two phenomena:
\begin{enumerate}
    \item Presenting ``If A then B'' before ``If B then C'' in the prompt generally achieves a higher accuracy compared to the reversed order.
    \item The performance gap is more significant when the number of premises increases.
\end{enumerate} Intuitively, such a  preference on the premise order aligns with human preference~\citep{dekeyser2000preferred} because in the preferred order, each derivation step can be done on-the-fly while looking at premises one by one, without needing to look back and forth across all premises at each step.

We conduct a systematic study on the premise order effect using a variety of SoTA LLMs, including GPT-4-turbo, GPT-3.5-turbo~\citep{achiam2023gpt}, PaLM 2-L~\citep{anil2023palm}, and Gemini 1.0 Pro~\citep{team2023gemini}. Our primary focus is deductive reasoning, and we benchmark all LLMs on problems that only involve \textit{modus ponens} (if P then Q; P; therefore Q), where all LLMs in our evaluation at least achieve decent performance with a small number of premises. We show that the accuracy decrease caused by different ordering can be more than 30\%. The ordering effect is further amplified when irrelevant premises (i.e., premises that are not needed to derive a conclusion) are presented in the prompt. Figure~\ref{fig:headline} illustrates a failure case, where all LLMs fail to generate the proof after changing the order of relevant rules. Interestingly, while all LLMs perform best when the premise order follows the ground truth proof, they reveal different preferences on other alternative orderings. Specifically, compared to randomly ordering the premises, GPT-4-turbo and GPT-3.5-turbo generally achieve better performance when the premise order is exactly the reverse of the ground truth proof, which enables LLMs to perform derivation via backward chaining. On the other hand, PaLM 2-L generally achieves the \textbf{worst} \textbf{performance} with such a reversed order.

Besides logical reasoning, we construct \gsmname{} to further investigate the ordering effect on mathematical reasoning. Specifically, we build \gsmname{} on top of a subset of GSM8K experiments, where we change the order of sentences in the problem description and manually verify that the ground truth answer remains the same. Our experiments again show that the performance of all LLMs notably drop, especially on longer problems that require more reasoning steps.

Our evaluation highlights that even in reasoning domains where the premise order \textbf{does not matter}, premise order \textbf{does matter in LLM reasoning}. Specifically, the premise ordering effect indicates that LLMs are more comfortable reasoning via reading left-to-right instead of back-and-forth, which can be attributed to the auto-regressive model design or the reasoning bias learned from the training corpus. We leave proposing new training and modeling techniques to mitigate the premise order effect as future work.

\section{Benchmarks}
\label{sec:benchmark}

\subsection{Logical Reasoning}
\label{sec:benchmark-logic}

Prior work has revealed the weaknesses of LLMs in logical reasoning~\citep{han2022folio,xu2023large,saparov2023testing,saparov2022language,wan2024b,yan2023concise}, especially when the proof is long and requires the knowledge of multiple deduction theorems. To isolate the effect of premise orders, we focus on a confined problem space adapted from SimpleLogic~\citep{zhang2022paradox}, which only includes propositional logic problems with definite clauses. Specifically, each problem includes: (1) a set of facts $A_1$,$\hdots$, $A_n$ that hold true; (2) a set of rules of the form ``If $X$, then $Y$'', ``If $X_0$ and $X_1$, then $Y$'', or ``If $X_0$ and $X_1$ and $X_2$, then $Y$''; and (3) a conclusion ``$C$ is \texttt{True}'' to be proved. As opposed to SimpleLogic --- which formulates the problem as a binary classification task (i.e., indicate whether the conclusion is \texttt{True} or \texttt{False}) --- in our benchmark, every problem has a ground-truth label of \texttt{True}, and we consider the prediction to be correct only when the generated proof is completely valid. With these strict criteria, the LLM is required to produce the step-by-step deduction that leads to the conclusion, and any hallucination of non-existent facts and rules is considered erroneous.

The key characteristic of our benchmark is that for each logical reasoning problem, we \textbf{synthetically generate variants} with \textbf{different premise orders.} Specifically, we denote the order that conforms to the ground truth proof with forward chaining as the \emph{forward} order, where the rule applied in each derivation step is sequentially presented in the problem description. Intuitively, presenting premises in the forward order simplifies the problem for humans, as this allows us to write the proof on-the-fly while reading the premises. Conversely, a premise ordering that is more random increases the task difficulty, since carrying out the derivation requires us to repetitively look for premises for each reasoning step. Motivated by this intuition, we categorize different premise orders based on their Kendall tau distance $\tau$~\citep{cicirello2019kendall,sen1968estimates} to the forward order, normalized into the range $[-1, 1]$. Specifically, $\tau=1$ is the \textit{forward} order, and we denote the order with $\tau=-1$ as the \emph{backward} order, which is the reverse of the forward order and aligns with the proof via backward chaining. $\tau$ $\approx 0$ suggests that there is no strong correlation between the premise order in the problem description and the proof. To thoroughly investigate the LLM preference on different premise orders, we evaluate the model performance on $\tau = 0.5$, $0$ and $-0.5$, in addition to the forward ($\tau=1$) and backward ($\tau=-1$) orders. We present examples with $\tau=1$ and $0$ in Figure~\ref{fig:headline}, and defer examples with other $\tau$ values to Figure~\ref{fig:ex-logical-order} in Appendix~\ref{app:logic-ex}.

We measure the premise order effect by varying the following two factors:

\begin{itemize}
    \item \textbf{Number of rules required in the proof.} It is expected that the premise order effect is more significant with more rules. For our benchmark, we generate problems whose numbers of rules range from 4 to 12.
    \item \textbf{Number of distracting rules} (i.e., rules that are not useful for the proof) presented in the problem. The presence of distracting rules also complicates the problem, as premise selection itself is challenging~\citep{wang2017premise,ferreira2020premise,irving2016deepmath}, and LLMs are shown to be easily distracted by irrelevant context~\citep{shi2023large}. We include problem variants with 0, 5 and 10 distracting rules.
\end{itemize}

We generate 200 problems for each number of required rules. Considering different premise orders and numbers of distracting rules, each problem includes 15 variants, resulting in a total of 27K problems in our benchmark.

\subsection{\gsmname{} for Mathematical Reasoning}
\label{sec:benchmark-math}

\textbf{\begin{figure*}[h!]
\includegraphics[width=1\linewidth]{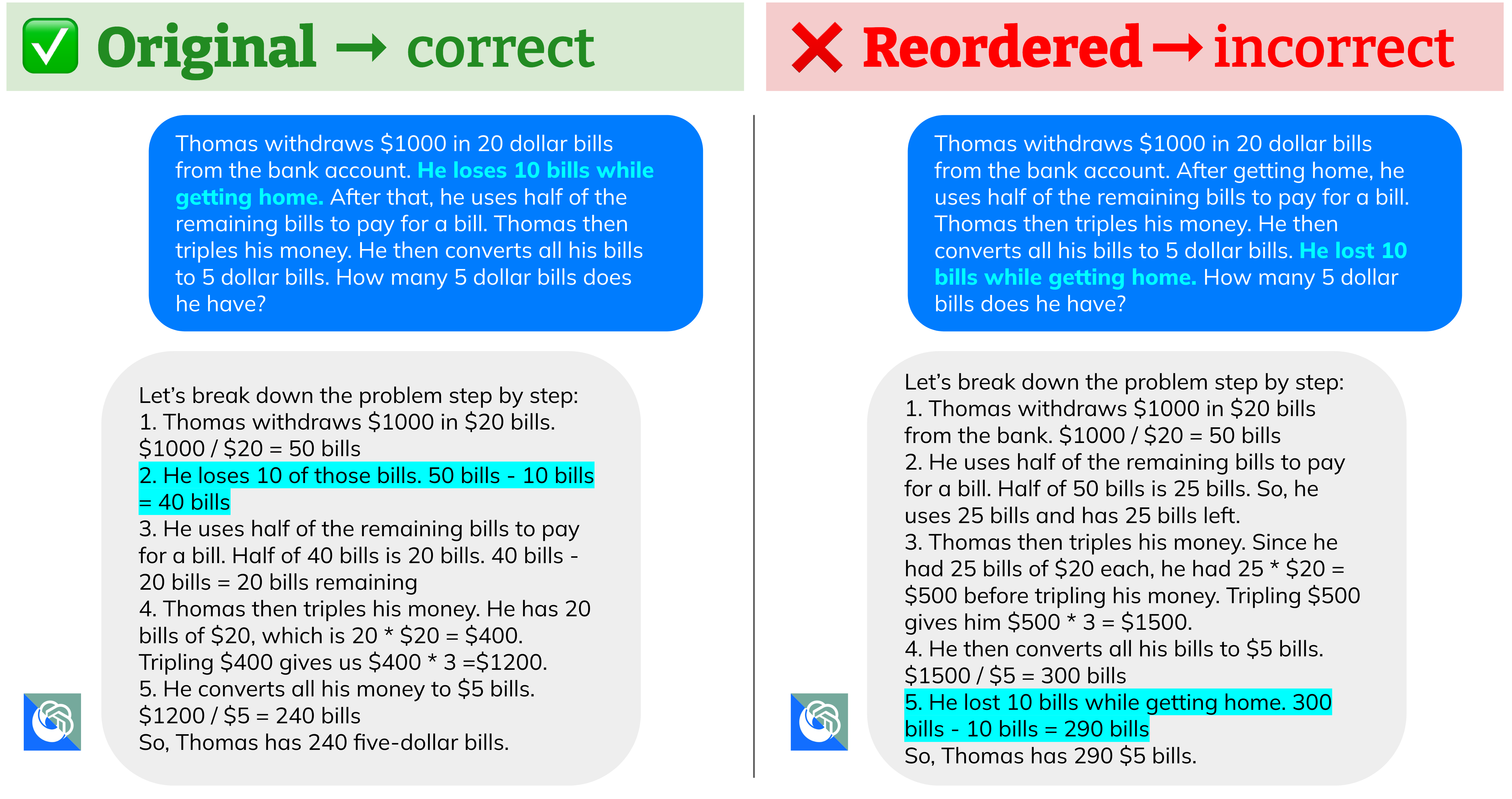}
\caption{\gsmname{} example where the original problem can be correctly solved by all LLMs in our evaluation, but all of them failed on the reordered one. Different calculation steps and their corresponding problem statements are annotated in light blue. Specifically, the reasoning steps of the original problem follows the ordering of problem statements, while the reordered problem does not.}
\label{fig:gsm-ex1}
\end{figure*}}

To further assess the effect of premise orders beyond logical reasoning, we construct the \gsmname{} dataset based on GSM8K~\citep{cobbe2021training}, which is a popular benchmark of grade school math word problems. Specifically, we first select GSM8K test problems with at least 5 sentences in the problem description, then filter out those problems where there is no alternative ordering that does not change the ground truth answer, e.g., problem statements that follow the causal order of an event series. For each of the remaining problem, we keep the last sentence untouched and rewrite the problem description with a different ordering of other sentences. Minor editing on words is allowed to ensure the grammatical correctness of the problem description. To facilitate the annotation process, for each problem, we write a simple function to enumerate all alternative orderings of problem statements until an ordering that causes the LLM prediction failure is discovered, which can be used for our manual rewriting if the alternative ordering found in the enumeration process happens to preserve the ground truth answer. In total, our \gsmname{} benchmark contains 220 pairs of problems, including both the original GSM8K problem description and the manually rewritten one with a different ordering of problem statements. Despite that over 60\% of problems in \gsmname{} only have 5 sentences, and all problems have at most 8 sentences, our evaluation shows that all LLMs still perform considerably worse on rewritten problems. Figure~\ref{fig:gsm-ex1} presents an example in \gsmname{} where all LLMs correctly solve the original problem but not the rewritten one. Specifically, the reasoning steps for the original problem follows the ordering of problem statements, while for the rewritten problem, the second calculation step in the correct solution should refer to the second-to-last sentence instead of the second sentence in the problem description. We provide a more detailed case study in Section~\ref{sec:exp-math}, and present the full dataset statistics in Appendix~\ref{app:gsm-stats}.
\section{Experiments}
\label{sec:exp}

\subsection{Experimental Setup}
\label{sec:setup}

We evaluate the premise ordering effect on GPT-4-turbo, GPT-3.5-turbo, PaLM 2-L and Gemini 1.0 Pro. We perform the greedy decoding with the temperature 0, and apply the zero-shot prompting in all experiments. On \gsmname{}, the model input only contains the problem description without additional instructions. For logical reasoning, as shown in Figure~\ref{fig:headline}, we add an instruction in the prompt to ask for a derivation that specifies which premise is used in each step.

\subsection{Logical Reasoning}
\label{sec:exp-logic}

\begin{figure*}[htbp!]
  \centering
  \includegraphics[width=\textwidth]{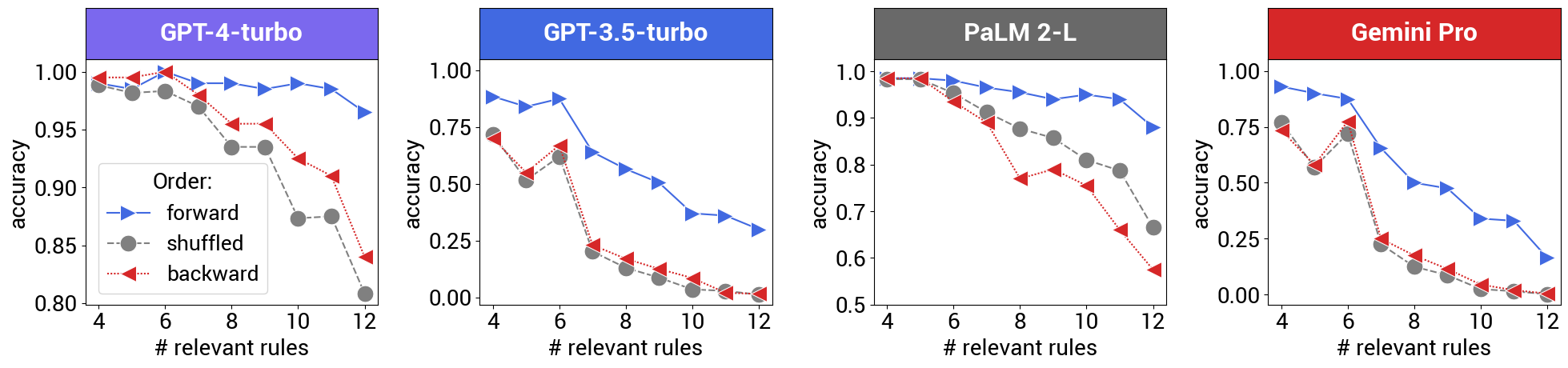}
  \caption{Logical reasoning without distracting rules. See Table~\ref{tab:logic-num-rules} in Appendix~\ref{app:exp-logic} for accuracy numbers.}
  \label{fig:logic-num-rules}
\end{figure*}

\begin{figure*}[htbp!]
  \centering
  \includegraphics[width=\textwidth]{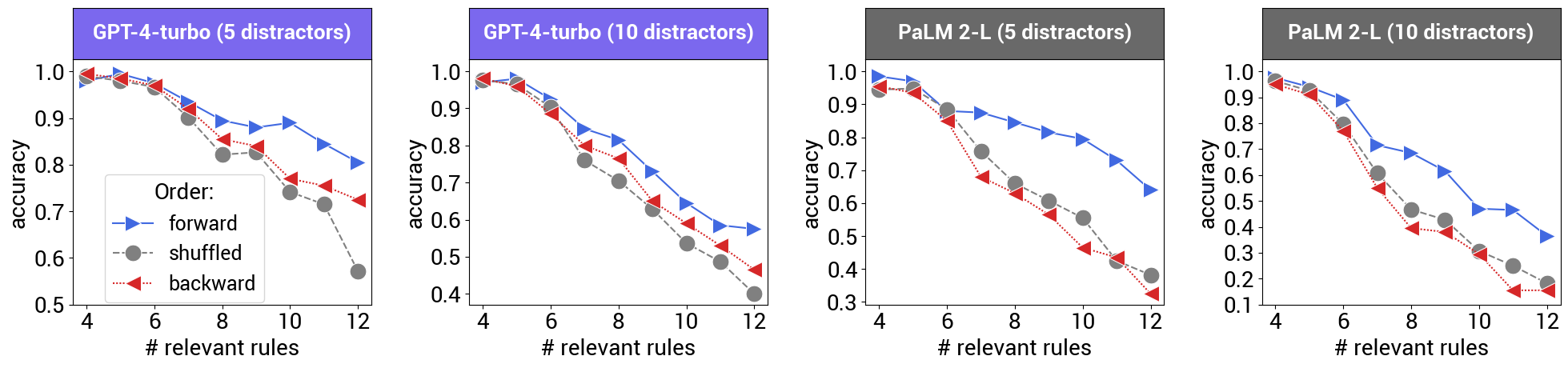}
  \caption{Logical reasoning with distracting rules. See Tables~\ref{tab:logic-distractors-5} and~\ref{tab:logic-distractors-10} for accuracy numbers.}
  \label{fig:logic-distractors}
\end{figure*}

\begin{figure*}[htbp!]
  \centering
  \includegraphics[width=\textwidth]{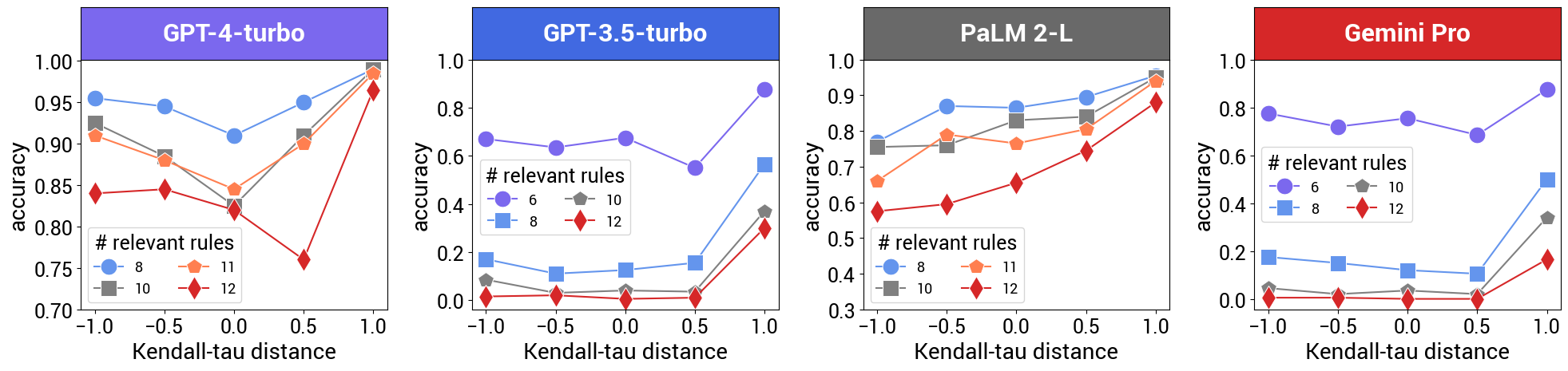}
  \caption{Results on different $\tau$ without distracting rules. See Table~\ref{tab:logic-order-0-distractors} for accuracy numbers.}
  \label{fig:logic-order-0-distractors}
\end{figure*}

\begin{figure*}[htbp!]
  \centering
  \includegraphics[width=\textwidth]{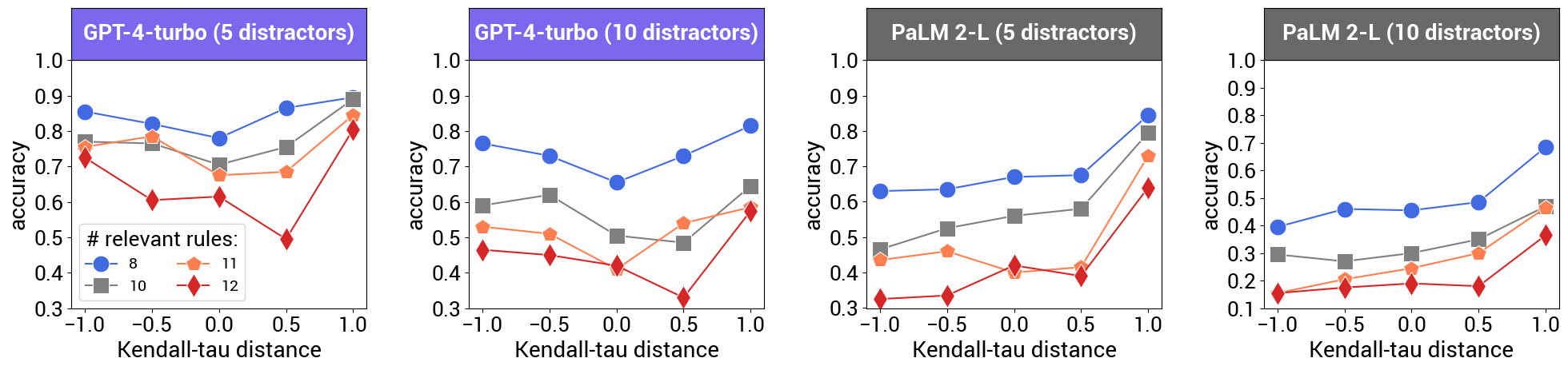}
  \caption{Results on different $\tau$ with distracting rules. See Tables~\ref{tab:logic-order-5-distractors} and~\ref{tab:logic-order-10-distractors} for accuracy numbers.}
  \label{fig:logic-order-distractors}
\end{figure*}

Figure~\ref{fig:logic-num-rules} presents the results with different numbers of relevant rules included in ground truth proofs, where the problem does not contain distracting rules, and the shuffled accuracy is the aggregation of results with $\tau$ = 0.5, 0 and -0.5. Across different LLMs, the forward order consistently achieves the best performance, which aligns with the human preference. The performance drop caused by alternative orderings becomes more significant when the number of rules increases. Meanwhile, models with weaker reasoning capabilities are also more sensitive to different premise orders. Specifically, while the accuracy decrease of GPT-4-turbo and PaLM 2-L is up to $20-30\%$, with Gemini 1.0 Pro and GPT-3.5-turbo, changing the premise order from the forward order can degrade the accuracy from over $65\%$ to below $25\%$, with an accuracy decrease of more than $40\%$.

\textbf{Breakdown on different premise orders.} We present the results of fine-grained breakdown on premise ordering in Figure~\ref{fig:logic-order-0-distractors}, where the orders are categorized based on Kendall tau distance $\tau$ as described in Section~\ref{sec:benchmark-logic}. Interestingly, while the top preference of all LLMs is the forward order, their preferences on other orders are not alike. Specifically, GPT-4-turbo generally prefers the backward order over other orders, and the overall performance decreases with a smaller absolute value of $\tau$. This observation is also consistent with the human reasoning pattern, as backward chaining is another well-established inference method. On the other hand, PaLM 2-L generally performs the worst with the backward order. With the decrease of $\tau$ (i.e., the premise order deviates more from the forward order), the accuracy drops. The preferences of Gemini 1.0 Pro and GPT-3.5-turbo are less consistent, still they prefer the backward order more often than other non-forward premise orders.

\textbf{Effect of distracting rules.} We assess the effect of distracting rules of GPT-4-turbo and PaLM 2-L, which reach a decent performance without the presence of distracting rules. Figures~\ref{fig:logic-distractors} and~\ref{fig:logic-order-distractors} show that adding distracting rules further decreases the reasoning performance and magnifies the effect of different premise orders. Still, the overall preferences of both LLMs remain the same as the scenario without distracting rules. Specifically, both LLMs again achieve the best performance with the forward order, and GPT-4-turbo prefers the backward order over other non-forward orders, while PaLM 2-L performance decreases with a smaller $\tau$.

\textbf{Error analysis.} In Table~\ref{tab:logic-errors}, we present the breakdown on prediction errors with different premise orders. We consider the following error categories:

\begin{enumerate}
\item \emph{wrong refutation}: the LLM wrongly claims that the conclusion can not be proved;
\item \emph{rule hallucination}: the LLM generates rules that do not exist in the problem;
\item \emph{fact hallucination}: the LLM generates facts that do not exist in the problem and are unproven.
\end{enumerate}

We observe that for all LLMs, fact hallucination is typically the most common error pattern, and this error type escalates dramatically with the decrease of $\tau$. The main reason is that LLMs are inclined to use the rules in the sequential order as they present in the problem, so when the next rule in the problem is not yet applicable, LLMs might still hallucinate facts to complete the proof step. Simultaneously, we observe that the percentage of wrong refutation is generally lower for $\tau=-1$ than for $|\tau|<1$. We present an example of wrong refutation in Figure~\ref{fig:headline}, and we include more examples of rule and fact hallucination in Figure~\ref{fig:logic-ex7} of Appendix~\ref{app:logic-ex}.

\begin{table}[htbp!]
\centering
\scalebox{0.85}{
\begin{tabular}{lrrrrr}
\toprule
& $\tau$ & Correct & Wrong & \multicolumn{2}{c}{Hallucination} \\
       &     &         & Refutation   &            Rule & Fact  \\
\midrule
\multirow{5}{*}{GPT-4-turbo} &  1    &  96.5\%  &0.5\% &    1.5\%   &	1.5\%    \\
     & 0.5    & 76.0\%	&10.5\% &    2.0\%  &  11.5\%  \\
     & 0      & 82.0\%  &4.5\% &	3.5\%  &	 10.0\%    \\
     & -0.5   & 84.5\%  &1.0\% &  4.5\%  &   10.0\%  \\
     & -1     & 84.0\%  &0.0\% &	3.5\%  &  12.5\%  \\
\midrule
\multirow{5}{*}{GPT-3.5-turbo} &  1    &  30.0\% &24.5\%& 9.5\% & 35.5\%    \\
     & 0.5    & 1.0\% & 54.5\% & 9.5\% & 33.0\%    \\
     & 0      & 0.5\% & 55.0\% & 7.5\% & 34.5\%     \\
     & -0.5   & 2.0\% & 50.0\% & 8.5\% & 37.5\% \\
     & -1     & 1.5\% & 34.5\% & 14.5\% & 47.0\%  \\
\midrule
\multirow{5}{*}{PaLM 2-L} &  1   & 88.0\% & 0.5\% & 3.0\% & 8.5\% \\
     & 0.5    & 74.5\% & 1.5\% & 9.5\% & 14.5\% \\
     & 0      & 65.5\% & 2.0\% & 11.0\% & 21.5\% \\
     & -0.5   & 59.5\% & 1.5\% & 10.0\% & 29.0\% \\
     & -1     & 57.5\% & 1.0\% & 11.5\% & 30.0\% \\     
\midrule
\multirow{5}{*}{Gemini 1.0 Pro} &  1    &  16.5\% & 28.0\% & 5.0\% & 50.5\% \\
     & 0.5    & 0.0\% & 59.0\% & 3.5\% & 37.5\% \\
     & 0      & 0.0\% & 34.0\% & 9.0\% & 57.0\% \\
     & -0.5   & 0.5\% & 24.5\% & 9.5\% & 65.5\% \\
     & -1     & 0.5\% & 27.5\% & 11.5\% & 60.5\% \\
\bottomrule
\end{tabular}}
\caption{Error analysis for logical reasoning with 12 relevant rules and no distracting rules.}
\label{tab:logic-errors}
\end{table}

\subsection{\gsmname{} for Mathematical Reasoning}
\label{sec:exp-math}

\begin{table}[htbp!]
\centering
\begin{subtable}{0.49\linewidth}
\begin{tabular}{lcc}
\toprule
 & Init Acc & Reorder Acc  \\
\midrule
GPT-4-turbo & 94.1\% & 85.0\% \\
PaLM 2-L & 86.4\% & 79.5\% \\
Gemini 1.0 Pro & 80.5\% & 69.1\% \\
GPT-3.5-turbo & 67.3\% & 51.8\% \\
\bottomrule
\end{tabular}
\caption{}
\label{tab:gsm-res}
\end{subtable}
\begin{subtable}{0.49\linewidth}
\centering
\begin{tabular}{lcc}
\toprule
 & Init Acc & Reorder Acc \\
\midrule
GPT-4-turbo & 100\% & 89.9\% \\
PaLM 2-L & 100\% & 87.9\% \\
Gemini 1.0 Pro & 100\% & 74.6\% \\
GPT-3.5-turbo & 100\% & 64.9\% \\
\bottomrule
\end{tabular}
\caption{}
\label{tab:gsm-res-sub}
\end{subtable}
\caption{Results on the \gsmname{} dataset: (a) accuracies on the full dataset; (b) for each model, the accuracies on the \gsmname{} subset where the original problems are correctly solved, thus the initial accuracy is 100\% for all models.}
\end{table}

\begin{figure*}[htbp!]
  \centering
  \includegraphics[width=\textwidth]{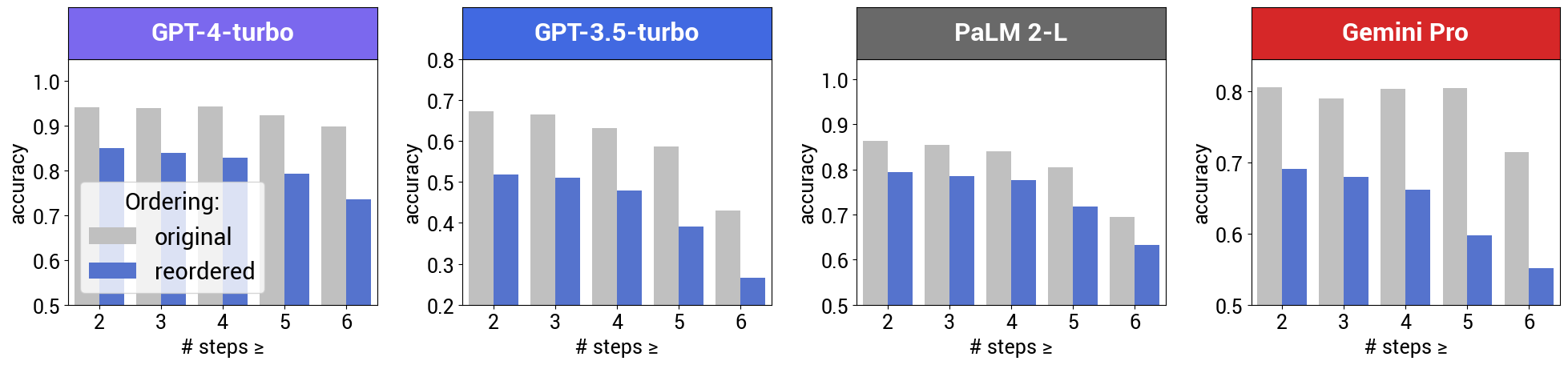}
  \caption{\gsmname{} results with different numbers of reasoning steps in the ground truth. See Table~\ref{tab:gsm-res-step} in Appendix~\ref{app:exp-gsm} for accuracy numbers.}
  \label{fig:gsm-step}
\end{figure*}

\begin{figure*}[htbp!]
  \centering
  \includegraphics[width=\textwidth]{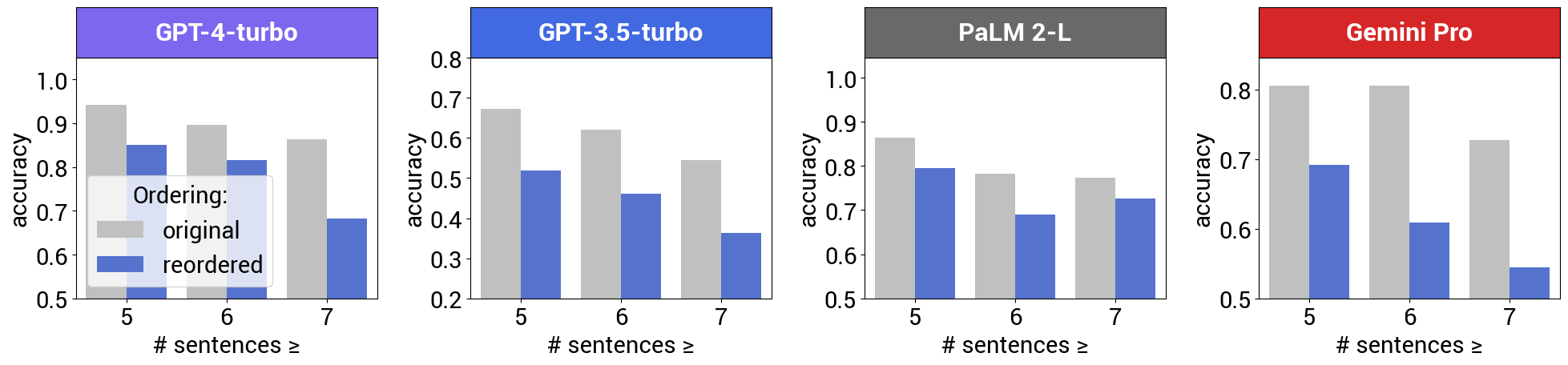}
  \caption{\gsmname{} results with different problem lengths. See Table~\ref{tab:gsm-res-len} for accuracy numbers.}
  \label{fig:gsm-len}
\end{figure*}

\begin{table}[h]
\centering
\begin{tabular}{lccc}
\toprule
 & Temporal & Unknown & Others  \\
\midrule
GPT-4-turbo & 45.0\% & 15.0\% & 40.0\%  \\
GPT-3.5-turbo & 21.6\% & 19.6\% & 58.8\% \\
PaLM 2-L & 34.8\% & 4.3\% & 60.9\% \\
Gemini 1.0 Pro & 29.5\% & 18.2\% & 52.3\% \\
\bottomrule
\end{tabular}
\caption{Error analysis on \gsmname{}. ``Temporal'' refers to the temporal order, and ``Unknown'' refers to the unknown variables.}
\label{tab:gsm-error-analysis}
\end{table}

\begin{figure*}[t]
\includegraphics[width=1\linewidth]{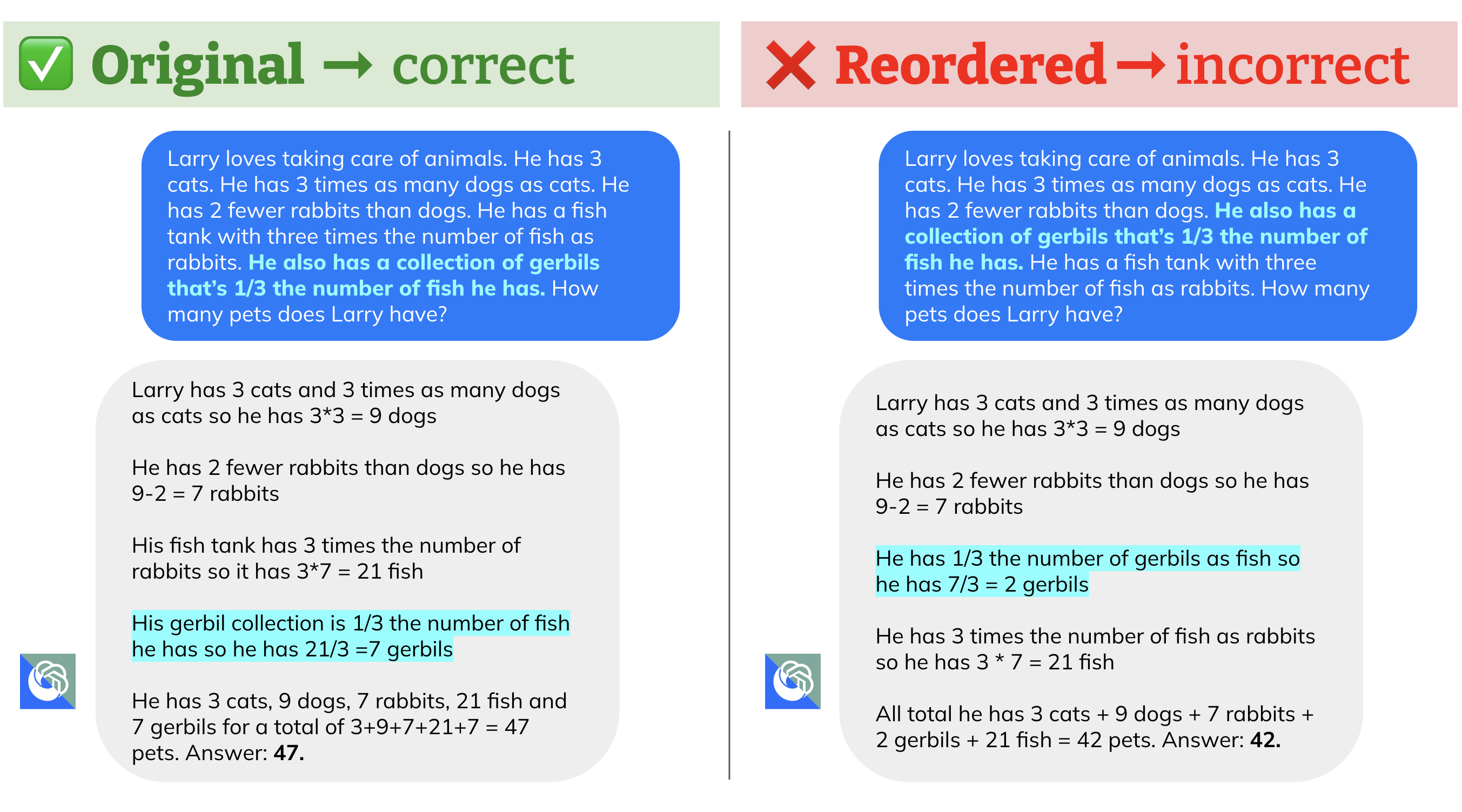}
\caption{\gsmname{} example where the original problem can be correctly solved by all LLMs, but GPT-3.5-Turbo fails on the reordered version while all the other LLMs still solve it correctly.}
\label{fig:gsm-ex2}
\end{figure*}

Table~\ref{tab:gsm-res} demonstrates the overall results on \gsmname{}. Again, all LLMs achieve a lower performance on \gsmname{}. Note that the original GSM8K problems are not necessarily written in the most preferable way, and thus sometimes the manual rewriting facilitates the reasoning and allows the model to correctly solve the reordered version of a problem that it fails on the original one. Therefore, in Table~\ref{tab:gsm-res-sub}, for each LLM, we also present the accuracy on those problems with their original descriptions solved by the model. We show that all LLMs fail on at least 10\% of reordered problems that they are initially able to solve, and this performance degradation is more than 35\% with GPT-3.5-turbo.

\textbf{Breakdown of problem complexity.} Figures~\ref{fig:gsm-step} and~\ref{fig:gsm-len} present the breakdown results on different number of reasoning steps and different number of problem sentences, respectively. Unsurprisingly, across all LLMs, the proof accuracy suffers on problems that require more reasoning steps and contain a greater number of sentences. Overall, the gap between the accuracies on initial and rewritten problems is more significant with more reasoning steps and longer problems for both GPT-4-turbo and Gemini 1.0 Pro, while the gap remains similar across different numbers of reasoning steps and problem lengths for PaLM 2-L and GPT-3.5-turbo.

\textbf{Error analysis.} To further understand the failure modes, for each LLM, we analyze those error cases where the original problems can be correctly solved but not the reordered ones, and we categorize the common error types in Table~\ref{tab:gsm-error-analysis}. Similar to our observation in logical reasoning experiments, the prediction errors in \gsmname{} are primarily due to the LLMs blindly using numbers in the sequential order of their appearances in the problem. Specifically, the most common error case for all LLMs is their tendency to overlook temporal order. Figure~\ref{fig:gsm-ex1} presents such an example, where the prediction failure is because some earlier events are described in the later part of the problem. Another category of errors occurs when some quantities are not specified while processing the problem in the sequential order, which introduces unknown variables for calculation. Take, for example, the problem in Figure~\ref{fig:gsm-ex2}. In the original problem, the number of each animal can be directly calculated based on its preceding sentence. However, in the reordered problem, the number of gerbils cannot directly be computed based on the preceding sentences, since the number of fish remains unknown up to that point, and the LLM must read the remaining sentences and calculate the number of fish first. However, the prediction from GPT-3.5-turbo instead uses the number calculated in the previous step (i.e., the number of rabbits) to calculate the number of gerbils, resulting in an error. Such a failure mode is less common with PaLM 2-L, but still constitutes a non-negligible proportion of prediction errors for the other LLMs. We present more examples of model predictions in Appendix~\ref{app:gsm-ex}.
\section{Related Work}
\label{sec:work}

\textbf{Failure modes of LLMs.} The premise order effect in this work is connected to several failure modes of LLMs in the literature, including the reversal curse~\citep{berglund2023reversal}, distractibility~\citep{shi2023large}, position bias~\citep{liu2024lost,wang2023large}, and limited capability of logical reasoning~\citep{han2022folio,xu2023large,saparov2023testing,saparov2022language,wan2024b,zhu2023large,yan2023concise}.
Specifically, \citet{shi2023large} show that including irrelevant context in the problem statement leads to a considerable performance drop on GSM8K and other reasoning benchmarks, revealing that LLMs are \textit{distractible}. This finding is in-line with our evaluation on logical reasoning, where we observe that adding irrelevant rules not only degrades the overall logical reasoning performance, but also escalates the premise order effect. The \textit{Reversal Curse}~\citep{berglund2023reversal} unveils another perspective of the order effect, where they show that an LLM that recognizes ``A is B'' does not necessarily learn that ``B is A.'' While their work studies the order effect between two entities within a single factual statement, our work focuses on reasoning problems with multiple premises, without restrictions on the number of (or relationship between) entities. In particular, for logical reasoning, we demonstrate that random permutations of premises often result in \textbf{worse} accuracy than the purely backward order. \citet{liu2024lost} discover the lost-in-the-middle phenomenon in the long-context scenario: the LLM performance is the best when the relevant information to solve the task is placed at the beginning or the end of the input context, while the performance is the worst when the LLM needs to utilize input context in the middle. In Appendix~\ref{app:lost-in-the-middle}, we show that lost-in-the-middle phenomenon does not affect the performance on our tasks, since the length of input problems does not exceed 300 tokens in our benchmark, which is relatively small compared to the context length limit of LLMs in our evaluation. \citet{yan2023concise} present an approach called Concise and Organized Perception for deductive reasoning, which first generates directed graphs by connecting facts and rules in the problem, then prune and reorder the context accordingly before calling the LLM to solve the problem. The improvement achieved by this approach again demonstrates the effect of premise ordering and irrelevant premises on logical reasoning. While such input preprocessing methods can mitigate the ordering effect on certain reasoning tasks, they require task-specific design and do not generalize across domains. We consider developing generic end-to-end reasoning techniques for LLMs to address the premise order effect as future work.

\textbf{Order effect for human logical reasoning.} Although the premise order does not matter in deductive reasoning, several studies show that the premise order can impact the human reasoning performance~\citep{dekeyser2000preferred,girotto1997effect}. \citet{dekeyser2000preferred} described \emph{co-reference} as a human preference of premise order; i.e., humans prefer the premises to be presented in an order where they can draw immediate conclusions after seeing each one. In this work, we show that LLMs also have such a preference, and they achieve the best performance when the ordering of rules follows the ground truth proof.  \citet{girotto1997effect} studied how the premise order affects logical reasoning for humans, and found that the premise order has a significant effect in solving \textit{modus tollens} problems (i.e., if P, then Q; not Q; therefore, not P), but not \textit{modus ponens} problems (i.e., if P, then Q; P; therefore, Q). However, differing from our work, they studied the influence of different ordering between rules and facts, e.g., their experiments on \textit{modus tollens} problems show that presenting negation statements (not Q) before rules (if P, then Q) improves the performance over the reverse order. On the other hand, our work focuses on \textit{modus ponens} problems that are easier for both humans and LLMs, and we show that the LLM performance is still quite sensitive to the ordering of the premises.

\textbf{Order effect of language models.} Some prior works show that language models are able to understand permuted texts to some extent, i.e., after a random permutation of words, models usually preserve a reasonable performance~\citep{abdou2022word,sinha2020unnatural}. Moreover, ~\citet{cao2023unnatural} show that even when a large fraction of words are scrambled, GPT-4 still achieves decent performance on several reasoning benchmarks. In contrast to permuted texts in these works that are typically unnatural and nonsensical, our premise order permutations do not alter the semantic meaning and remain syntactically valid (we manually verify this). Nevertheless, we demonstrate that LLM reasoning performance is highly brittle to the ordering of the premises. For long-digit addition, prior works demonstrate that reversing the input numbers is a key to achieve better length generalization performance~\citep{lee2023teaching,zhou2023algorithms,zhou2024transformers}. Specifically, by reversing the input numbers so that the least significant digit is presented first, the Transformer learns a simpler way of performing addition, where the model only needs to perform computation with the corresponding digits of operands and the carry-on digit at each step, without the need of looking at other digits. This approach enables the Transformer to better perform addition when trained from scratch, which also aligns with our finding: after reversing the input numbers, the premise order (i.e., orders of digits) follows the right ordering of performing long-digit addition, thus enables Transformers to better learn the task.
\section{Conclusion}
\label{sec:conc}
In this work, we show that the premise order significantly affects LLMs' performance on reasoning tasks, even when the premise order does not change the underlying task itself. Our comprehensive evaluation demonstrates that LLM tendencies resemble human preference w.r.t. premise order, i.e., LLMs achieve the best performance when the premise order follows the intermediate reasoning steps to solve the problem. Conversely, LLMs face difficulties when the reasoning problem requires the model to read the problem description back-and-forth, resulting in a performance drop of over 30\%. We further extend the study to mathematical reasoning and present the \gsmname{} benchmark, and again experimentally confirm the ordering effect.

While humans also have a preference of premise orders for reasoning problems, LLMs are much more susceptible to such ordering effects. We can attempt to ascribe the premise order effect to several candidate factors, such as the auto-regressive model design, training objectives, and training data mixture. However, we leave proposing theoretical explanations of this limitation and developing new techniques towards addressing the premise order effect as future work.

\section*{Acknowledgment}

We would like to thank Chen Liang and Dale Schuurmans for helpful discussion and feedback.

\bibliography{ref}

\appendix
\onecolumn
\section{\gsmname{} Dataset Statistics}
\label{app:gsm-stats}

Table~\ref{tab:gsm-stat} presents the statistics of our \gsmname{} benchmark. 

\begin{table}[h]
\centering
\begin{subtable}{\columnwidth}
\centering
\begin{tabular}{lc}
\toprule
\# Steps & \# Problems \\
\midrule
2 & 20 \\
3 & 43 \\
4 & 65 \\
5 & 43 \\
6 & 23 \\
7 & 15 \\
8 & 11 \\
\bottomrule
\end{tabular}
\caption{}
\label{tab:gsm-stat-step}
\end{subtable}
\begin{subtable}{\columnwidth}
\centering
\begin{tabular}{lc}
\toprule
\# Sentences & \# Problems \\
\midrule
5 & 133 \\
6 & 65 \\
7 & 19 \\
8 & 3 \\
\bottomrule
\end{tabular}
\caption{}
\label{tab:gsm-stat-len}
\end{subtable}
\caption{Statistics of the \gsmname{} dataset, with 220 problems in total: (a) breakdown on the number of reasoning steps; (b) breakdown on the number of sentences in the questions.}
\label{tab:gsm-stat}
\end{table}

\section{Logical Reasoning Examples}
\label{app:logic-ex}
\begin{figure*}[htbp!]
\includegraphics[width=1\linewidth]{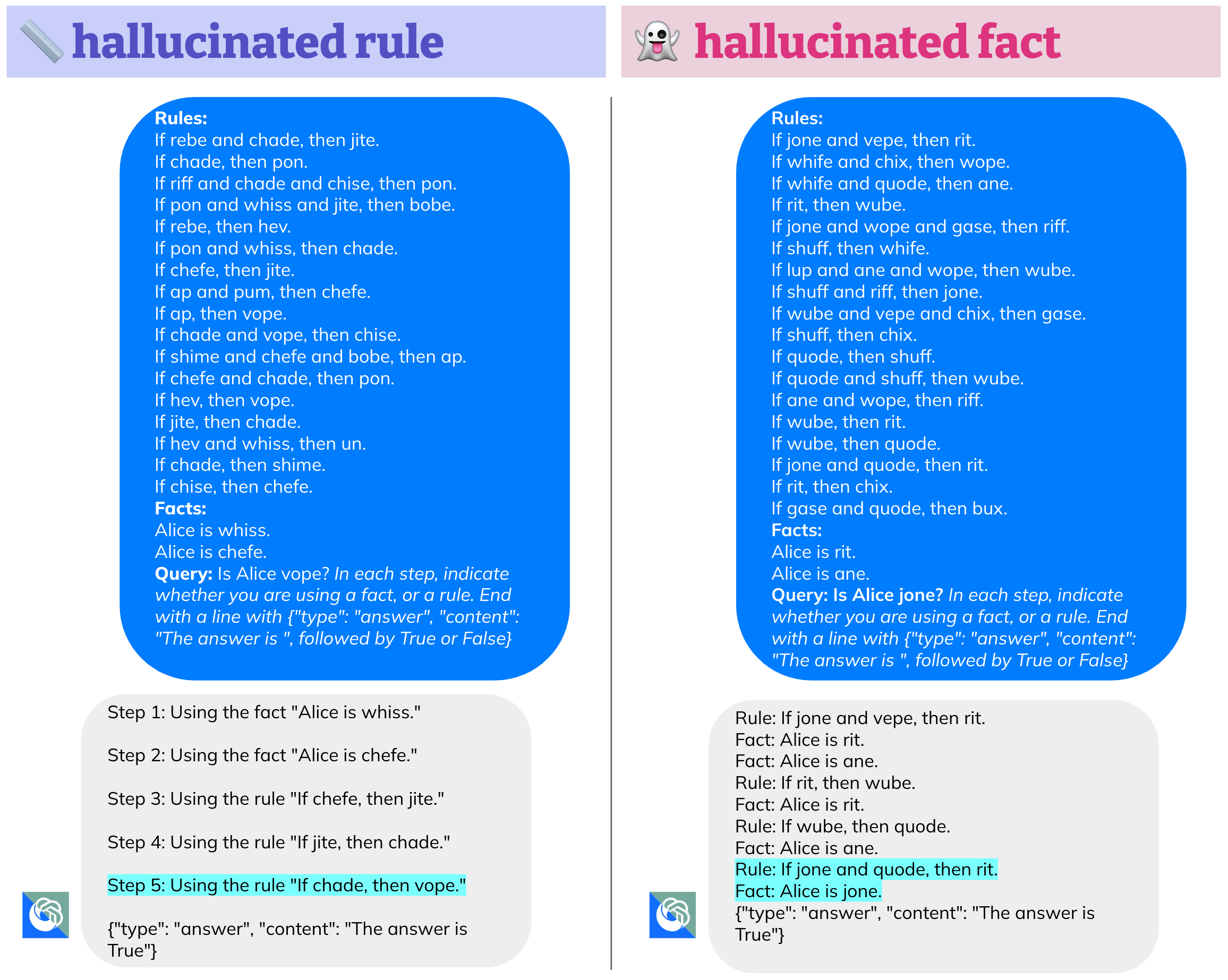}

\caption{Examples of hallucinated rules (left) and facts (right) produced by GPT-3.5-Turbo while solving our logical reasoning benchmark.}
\label{fig:logic-ex7}
\end{figure*}
Figure~\ref{fig:logic-ex7} presents common classes of errors --- hallucinated rules and facts --- by LLMs while solving our logical reasoning benchmark.

\begin{figure*}[t]
\includegraphics[width=1\linewidth]{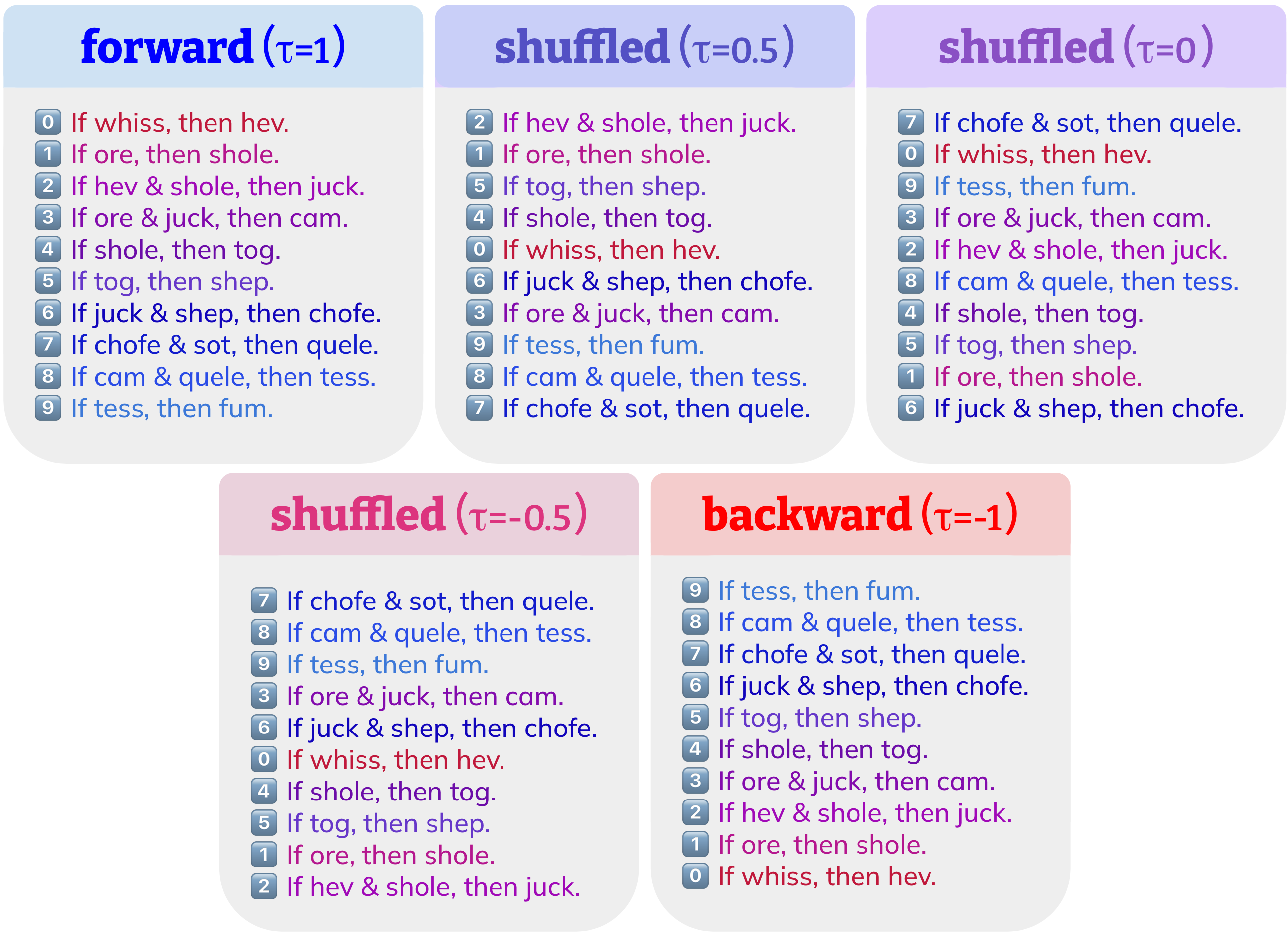}

\caption{An example logical reasoning problem with different premise orders. The number emojis are for ease of viewing. The ampersands were originally ``and''s in the original prompt. The facts and query have been excluded for brevity.}
\label{fig:ex-logical-order}
\end{figure*}

Figure~\ref{fig:ex-logical-order} presents a sample logical reasoning problem with premise orders of different $\tau$ values. We can see that the rules become less ordered when the absolute value of $\tau$ decreases.

\section{\gsmname{} Examples}
\label{app:gsm-ex}

In this section, we present more examples of LLM predictions on \gsmname{} problems.

\begin{figure*}[t]
\includegraphics[width=1\linewidth]{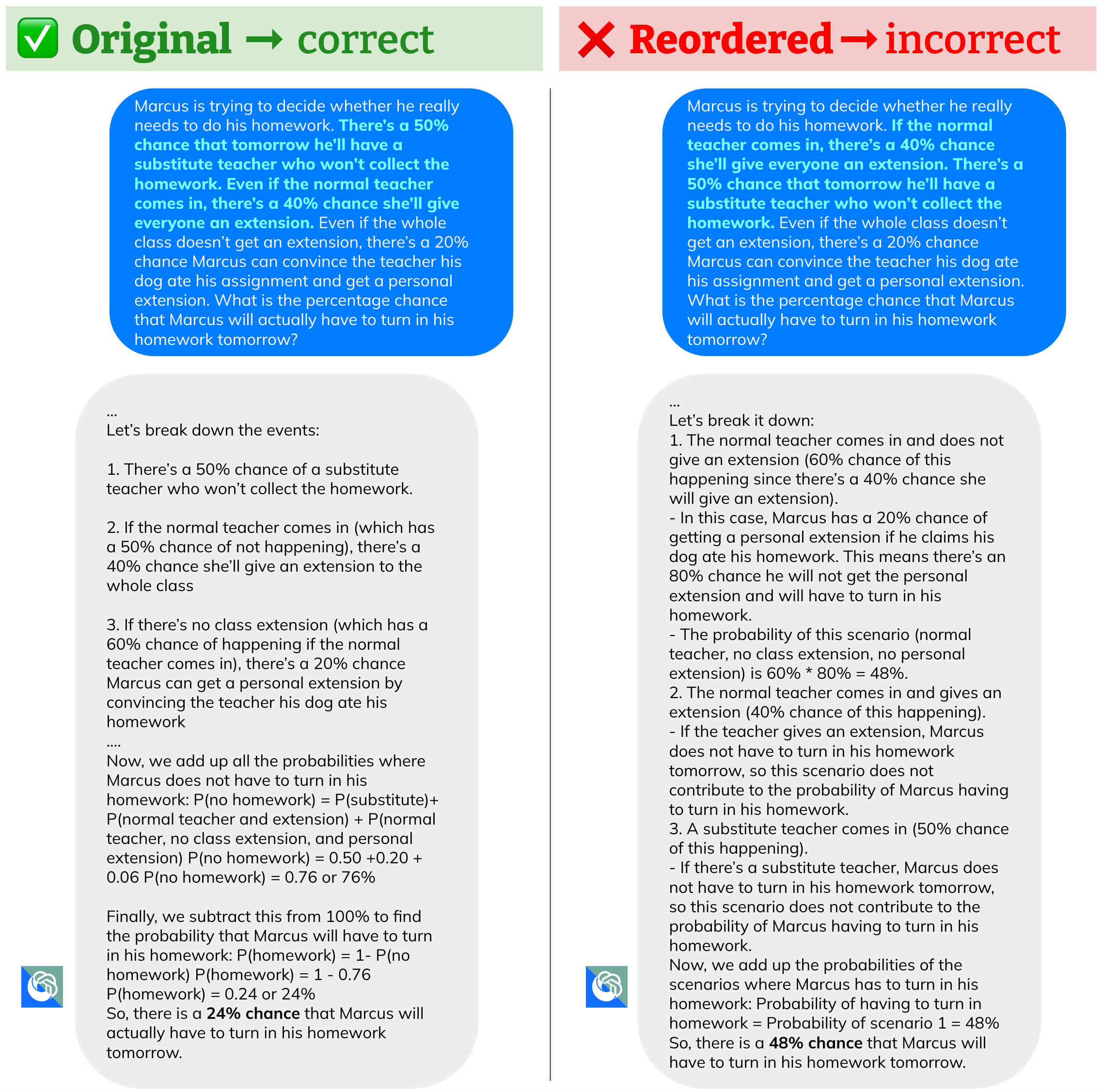}
\caption{\gsmname{} example where the original problem can be correctly solved by GPT-4 Turbo, but the model fails on the reordered one. }
\label{fig:gsm-ex3}
\end{figure*}

Figure~\ref{fig:gsm-ex3} presents a failure case of a probability problem, which falls into the ``Others'' category in the error analysis (Table~\ref{tab:gsm-error-analysis}). Specifically, in the reordered problem, after the LLM reads the sentence about the scenario with a normal teacher coming in, the LLM immediately attempts to compute the probability that Marcus has to turn in his homework, ignoring that the LLM needs to compute the probability that a normal teacher will come in using the next sentence.

\begin{figure*}[t]
\includegraphics[width=1\linewidth]{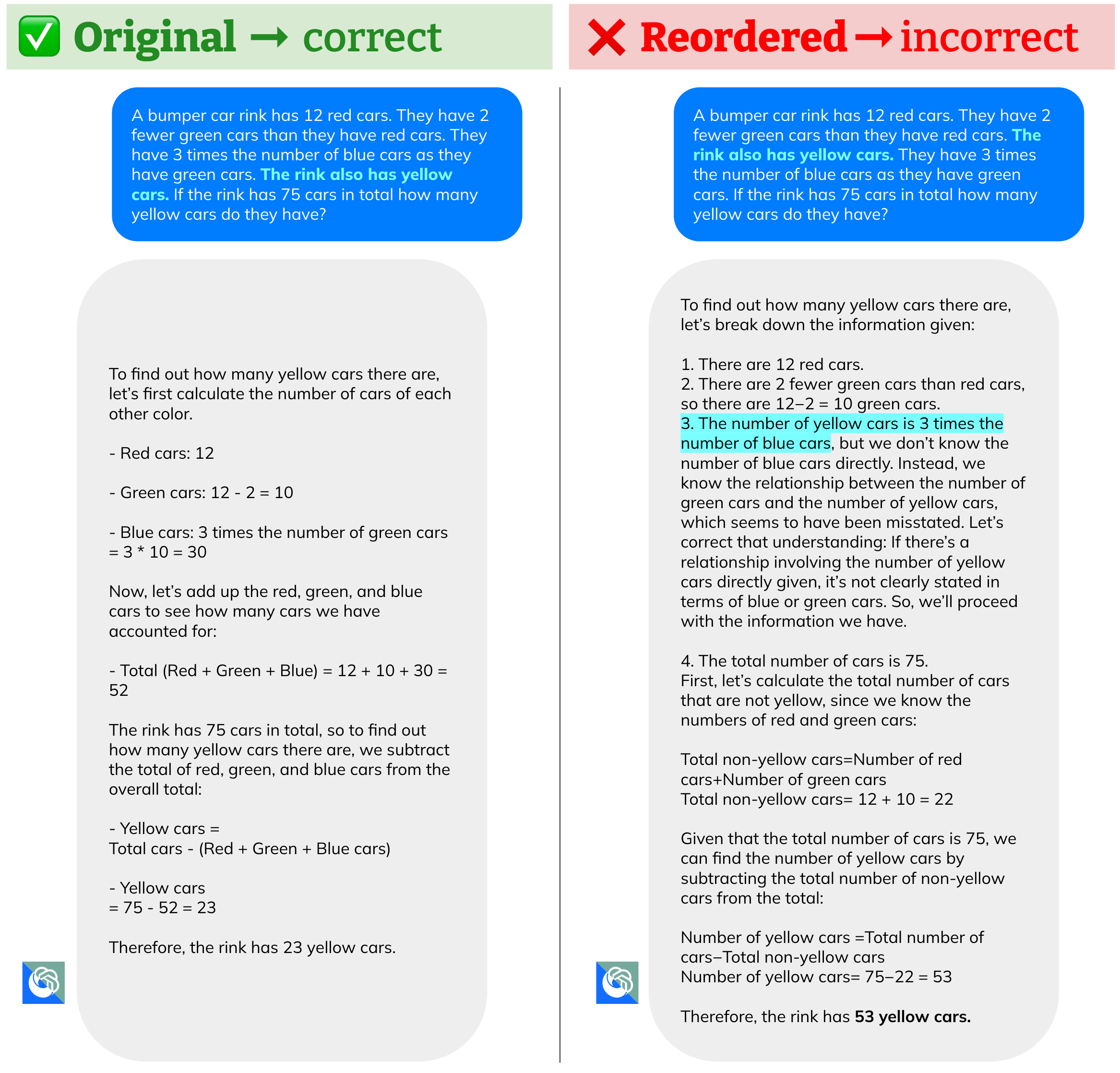}
\caption{\gsmname{} example where the original problem can be correctly solved by all models, but GPT-4 Turbo and Gemini Pro failed on the reordered one.}
\label{fig:gsm-ex4}
\end{figure*}

Figures~\ref{fig:gsm-ex4} shows another wrong prediction of GPT-4 Turbo, where the error pattern is analogous to rule hallucination in logical reasoning evaluation. Interestingly, when moving the sentence about yellow cars preceding to the sentence about quantities of blue and green cars, GPT-4 Turbo starts to hallucinate the relationship between the number of yellow cars and the number of blue cars, resulting in insufficient information to correctly solve the problem.

\begin{figure*}[t]
\includegraphics[width=1\linewidth]{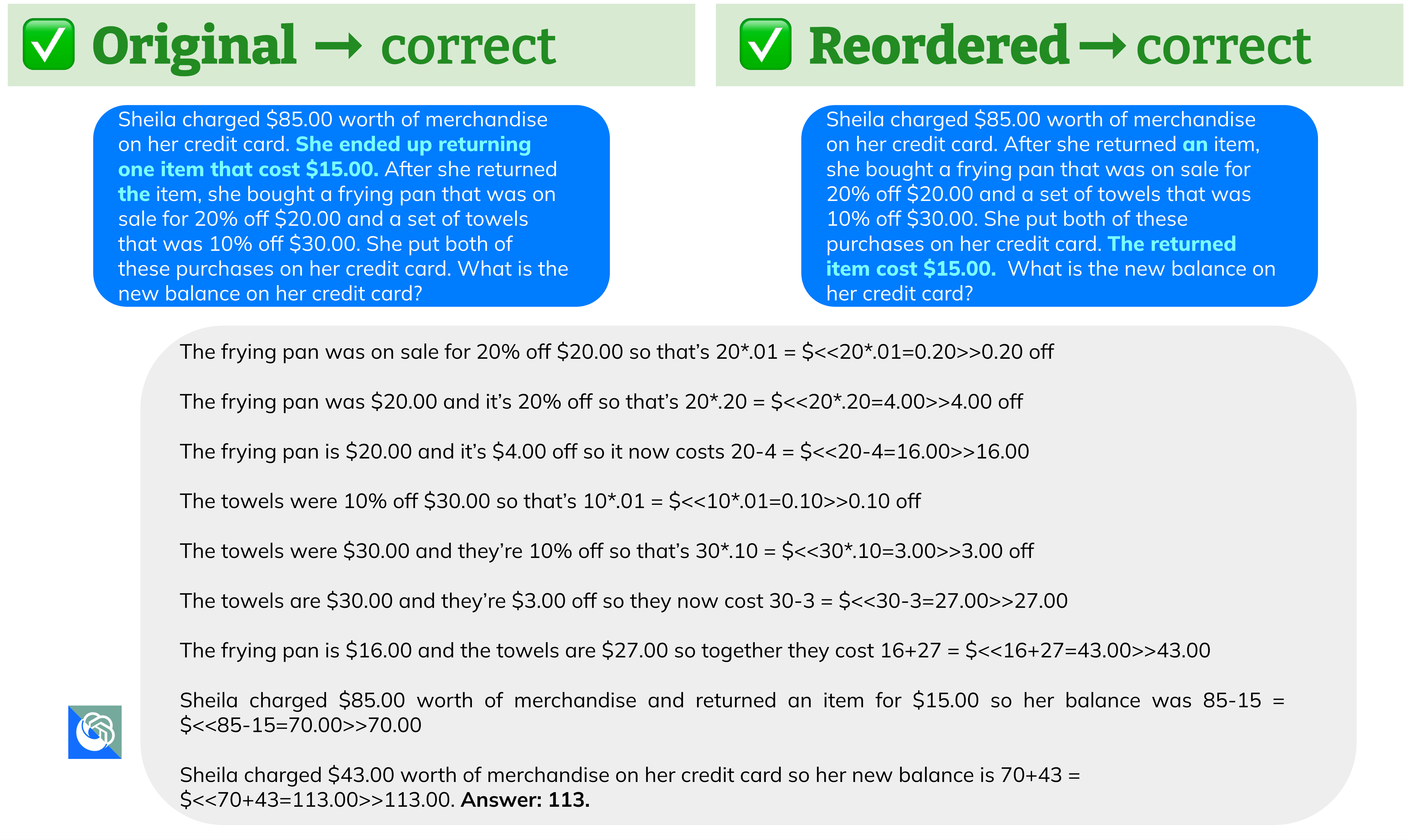}
\caption{\gsmname{} example where both the original and the reordered problems were correctly solved by all LLMs in our evaluation.}
\label{fig:gsm-ex5}
\end{figure*}

\begin{figure*}[htbp!]
\includegraphics[width=1\linewidth]{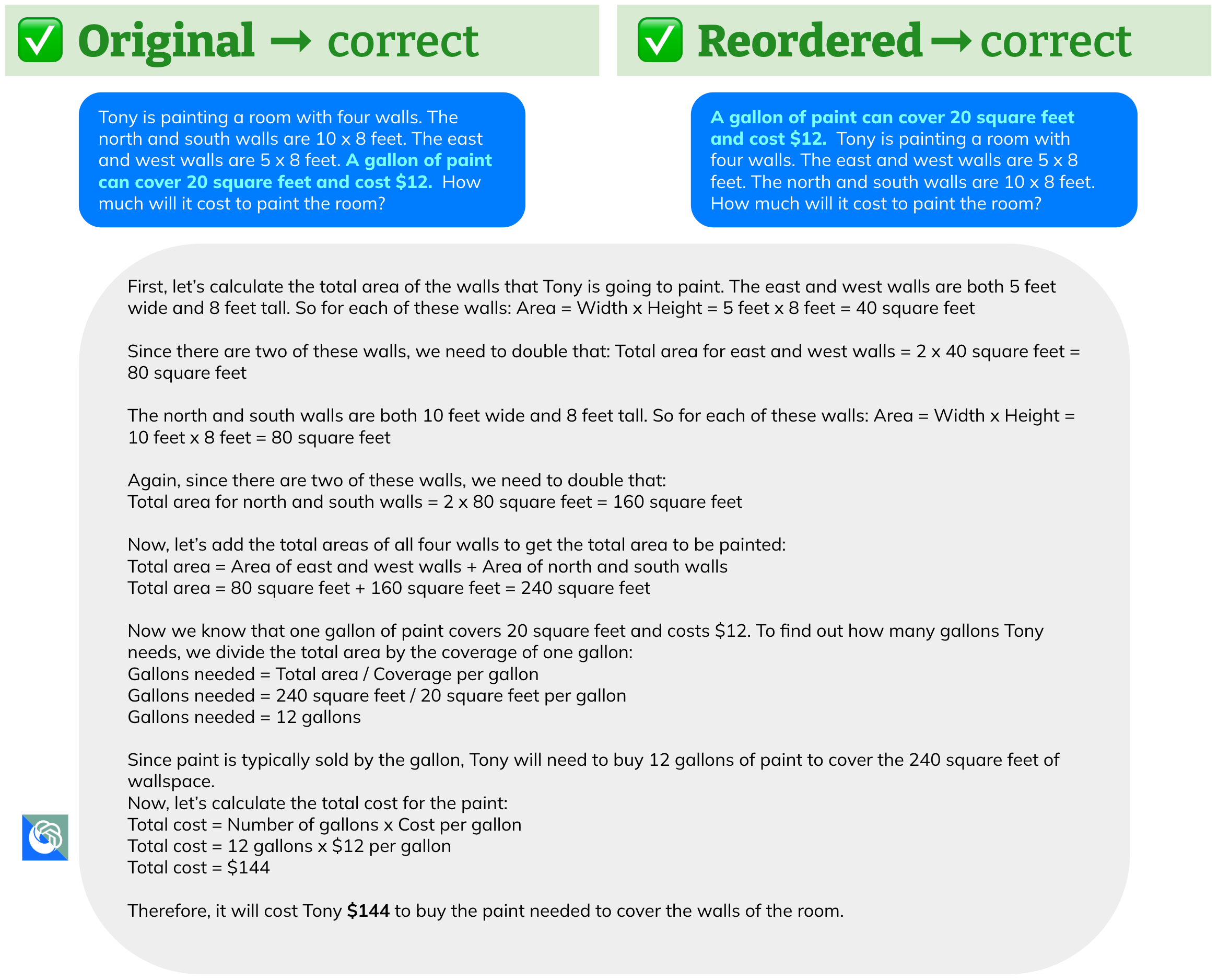}
\caption{\gsmname{} example where both the original and the reordered problems were correctly solved by all LLMs in our evaluation.}
\label{fig:gsm-ex6}
\end{figure*}

Figures~\ref{fig:gsm-ex5} and~\ref{fig:gsm-ex6} present examples where both the original and reordered problems are correctly solved by LLMs in our evaluation. In both original problems, the succeeding sentences do not strongly depend on the preceding sentences.


\section{Does Logical Reasoning Suffer from the Lost-in-the-middle Issue?}
\label{app:lost-in-the-middle}

\citet{liu2024lost} demonstrate that when the input context becomes long, LLMs might suffer from the lost-in-the-middle issue: the model performance significantly degrades when relevant information to solve the task is in the middle of the input, instead of at the beginning or the end. Therefore, when given distracting rules for logical reasoning, another potential factor that might affect the model performance is the position of relevant rules in the model input.

To examine the effect of such position bias, we conduct ablations on PaLM 2-L with 10 distracting rules, and we compare the performance with relevant rules added in the beginning, middle or the end of the problem description. Table~\ref{tab:lost-in-the-middle} shows that with the same order and number of rules, the variation in performance is very small, whereas changing the order significantly affects the results. Note that the longest inputs in our logical reasoning benchmark, i.e., problems with 12 relevant rules and 10 distracting rules, only contain no more than 300 tokens, which is relatively short compared to the context length limit of LLMs in our evaluation. These results confirm that on our tasks where the input problems (and thus input context) are short, lost-in-the-middle phenomenon is not the primary cause of the performance difference. In our primary experiments, for all logical reasoning problems, we interleave distracting rules with relevant rules in the input context.

\begin{table*}[h]
\centering
\begin{tabular}{lllll}
\toprule
\# rules, position \textbackslash order & Forward & Backward & Shuffled &  \\
\midrule
8, beginning                            & 68.0\%  & 40.0\%   & 45.5\%   &  \\
8, middle                               & 67.0\%  & 39.0\%   & 44.5\%   &  \\
8, end                                  & 67.0\%  & 40.0\%   & 45.5\%   &  \\
12, beginning                           & 36.5\%  & 17.0\%   & 16.0\%   &  \\
12, middle                              & 36.5\%  & 17.0\%   & 18.5\%   &  \\
12, end                                 & 35.0\%  & 16.0\%   & 19.5\%   & \\
\bottomrule
\end{tabular}
\caption{Logical reasoning results performance of PaLM 2-L, with relevant rules at different positions of the input context.}
\label{tab:lost-in-the-middle}
\end{table*}

\section{Full Results for Logical Reasoning}
\label{app:exp-logic}

Tables~\ref{tab:logic-num-rules} and~\ref{tab:logic-order-0-distractors} present the accuracy numbers for Figures~\ref{fig:logic-num-rules} and~\ref{fig:logic-order-0-distractors}, which are results on different numbers of relevant rules without distracting rules.

Tables~\ref{tab:logic-distractors-5} and~\ref{tab:logic-order-5-distractors} present the accuracy numbers for Figures~\ref{fig:logic-distractors} and~\ref{fig:logic-order-distractors} with 5 distracting rules.

Tables~\ref{tab:logic-distractors-10} and~\ref{tab:logic-order-10-distractors} present the accuracy numbers for Figures~\ref{fig:logic-distractors} and~\ref{fig:logic-order-distractors} with 10 distracting rules.

\begin{table*}[h]
\centering
\begin{subtable}{0.45\textwidth}
\centering
\scalebox{0.7}{
\begin{tabular}{llc}
\toprule
\# Rules & Order  & Acc  \\
\midrule
\multirow{3}{*}{4} & Forward & 99.0\% \\
& Backward & 99.5\%\\
& Shuffled & 98.8\%\\
\midrule
\multirow{3}{*}{5} & Forward & 98.5\%\\
& Backward & 99.5\%\\
& Shuffled & 98.2\%\\
\midrule
\multirow{3}{*}{6} & Forward & 100\%\\
& Backward & 100\%\\
& Shuffled & 98.3\%\\
\midrule
\multirow{3}{*}{7} & Forward & 99.0\%\\
& Backward & 98.0\%\\
& Shuffled & 97.0\%\\
\midrule
\multirow{3}{*}{8} & Forward & 99.0\%\\
& Backward & 95.5\%\\
& Shuffled & 93.5\%\\
\midrule
\multirow{3}{*}{9} & Forward & 98.5\%\\
& Backward & 95.5\%\\
& Shuffled & 93.5\%\\
\midrule
\multirow{3}{*}{10} & Forward & 99.0\%\\
& Backward & 92.5\%\\
& Shuffled & 87.3\%\\
\midrule
\multirow{3}{*}{11} & Forward & 98.5\%\\
& Backward & 91.0\%\\
& Shuffled & 87.5\%\\
\midrule
\multirow{3}{*}{12} & Forward & 96.5\%\\
& Backward & 84.0\%\\
& Shuffled & 80.8\%\\
\bottomrule
\end{tabular}}
\caption{GPT-4-turbo.}
\label{tab:logic-num-rules-gpt4}
\end{subtable}
\begin{subtable}{0.45\textwidth}
\centering
\scalebox{0.7}{
\begin{tabular}{llc}
\toprule
\# Rules & Order  & Acc  \\
\midrule
\multirow{3}{*}{4} & Forward & 98.5\% \\
& Backward & 98.5\%\\
& Shuffled & 98.3\%\\
\midrule
\multirow{3}{*}{5} & Forward & 98.5\%\\
& Backward & 98.5\%\\
& Shuffled & 98.3\%\\
\midrule
\multirow{3}{*}{6} & Forward & 98.0\%\\
& Backward & 93.5\%\\
& Shuffled & 95.3\%\\
\midrule
\multirow{3}{*}{7} & Forward & 96.5\%\\
& Backward & 89.0\%\\
& Shuffled & 91.2\%\\
\midrule
\multirow{3}{*}{8} & Forward & 95.5\%\\
& Backward & 77.0\%\\
& Shuffled & 87.7\%\\
\midrule
\multirow{3}{*}{9} & Forward & 94.0\%\\
& Backward & 79.0\%\\
& Shuffled & 85.7\%\\
\midrule
\multirow{3}{*}{10} & Forward & 95.0\%\\
& Backward & 75.5\%\\
& Shuffled & 81.0\%\\
\midrule
\multirow{3}{*}{11} & Forward & 94.0\%\\
& Backward & 66.0\%\\
& Shuffled & 78.7\%\\
\midrule
\multirow{3}{*}{12} & Forward & 88.0\%\\
& Backward & 57.5\%\\
& Shuffled & 66.5\%\\
\bottomrule
\end{tabular}}
\caption{PaLM 2-L.}
\label{tab:logic-num-rules-palm}
\end{subtable}
\begin{subtable}{0.45\textwidth}
\centering
\scalebox{0.7}{
\begin{tabular}{llc}
\toprule
\# Rules & Order  & Acc  \\
\midrule
\multirow{3}{*}{4} & Forward & 93.0\%\\
& Backward & 73.5\%\\
& Shuffled & 77.0\%\\
\midrule
\multirow{3}{*}{5} & Forward & 90.0\% \\
& Backward & 58.0\%\\
& Shuffled & 57.0\%\\
\midrule
\multirow{3}{*}{6} & Forward & 87.5\%\\
& Backward & 77.5\%\\
& Shuffled & 72.0\%\\
\midrule
\multirow{3}{*}{7} & Forward & 65.5\%\\
& Backward & 25.0\%\\
& Shuffled & 22.5\%\\
\midrule
\multirow{3}{*}{8} & Forward & 50.0\%\\
& Backward & 17.5\%\\
& Shuffled & 12.5\%\\
\midrule
\multirow{3}{*}{9} & Forward & 47.5\%\\
& Backward & 11.5\%\\
& Shuffled & 8.7\%\\
\midrule
\multirow{3}{*}{10} & Forward & 34.0\%\\
& Backward & 4.5\%\\
& Shuffled & 2.5\%\\
\midrule
\multirow{3}{*}{11} & Forward & 33.0\%\\
& Backward & 2.0\%\\
& Shuffled & 1.5\%\\
\midrule
\multirow{3}{*}{12} & Forward & 16.5\%\\
& Backward & 0.5\%\\
& Shuffled & 0.2\%\\
\bottomrule
\end{tabular}}
\caption{Gemini 1.0 Pro.}
\label{tab:logic-num-rules-gemini}
\end{subtable}
\begin{subtable}{0.45\textwidth}
\centering
\scalebox{0.7}{
\begin{tabular}{llc}
\toprule
\# Rules & Order  & Acc  \\
\midrule
\multirow{3}{*}{4} & Forward & 88.5\% \\
& Backward & 70.0\% \\
& Shuffled & 71.8\% \\
\midrule
\multirow{3}{*}{5} & Forward & 84.0\% \\
& Backward & 55.0\%\\
& Shuffled & 51.7\% \\
\midrule
\multirow{3}{*}{6} & Forward & 87.5\% \\
& Backward & 67.0\% \\
& Shuffled & 62.0\% \\
\midrule
\multirow{3}{*}{7} & Forward & 64.0\% \\
& Backward & 23.0\% \\
& Shuffled & 20.2\% \\
\midrule
\multirow{3}{*}{8} & Forward & 56.5\% \\
& Backward & 15.5\% \\
& Shuffled & 13.0\% \\
\midrule
\multirow{3}{*}{9} & Forward & 50.5\% \\
& Backward & 9.5\%\\
& Shuffled & 8.7\%\\
\midrule
\multirow{3}{*}{10} & Forward & 37.0\%\\
& Backward & 3.5\% \\
& Shuffled & 3.5\%\\
\midrule
\multirow{3}{*}{11} & Forward & 36.0\% \\
& Backward & 1.0\% \\
& Shuffled & 2.8\%\\
\midrule
\multirow{3}{*}{12} & Forward & 30.0\%\\
& Backward & 1.0\%\\
& Shuffled & 1.2\%\\
\bottomrule
\end{tabular}}
\caption{GPT-3.5-turbo.}
\label{tab:logic-num-rules-gpt3.5}
\end{subtable}
\caption{Result table corresponding to Figure~\ref{fig:logic-num-rules}.}
\label{tab:logic-num-rules}
\end{table*}
\begin{table*}[h]
\centering
\begin{subtable}{0.45\textwidth}
\centering
\begin{tabular}{llc}
\toprule
\# Rules & Order  & Acc  \\
\midrule
\multirow{3}{*}{4} & Forward & 98.0\%\\
& Backward & 99.5\%\\
& Shuffled & 99.0\%\\
\midrule
\multirow{3}{*}{5} & Forward & 99.5\%\\
& Backward & 98.5\%\\
& Shuffled & 98.0\%\\
\midrule
\multirow{3}{*}{6} & Forward & 97.5\%\\
& Backward & 97.0\%\\
& Shuffled & 96.7\%\\
\midrule
\multirow{3}{*}{7} & Forward & 93.5\%\\
& Backward & 92.0\%\\
& Shuffled & 90.2\%\\
\midrule
\multirow{3}{*}{8} & Forward & 89.5\%\\
& Backward & 85.5\%\\
& Shuffled & 82.2\%\\
\midrule
\multirow{3}{*}{9} & Forward & 88.0\%\\
& Backward & 84.0\%\\
& Shuffled & 82.7\%\\
\midrule
\multirow{3}{*}{10} & Forward & 89.0\%\\
& Backward & 77.0\%\\
& Shuffled & 74.2\%\\
\midrule
\multirow{3}{*}{11} & Forward & 84.5\%\\
& Backward & 75.5\%\\
& Shuffled & 71.5\%\\
\bottomrule
\multirow{3}{*}{12} & Forward & 80.5\%\\
& Backward & 72.5\%\\
& Shuffled & 57.2\%\\
\bottomrule
\end{tabular}
\caption{GPT-4-turbo.}
\label{tab:logic-distractors-5-gpt4}
\end{subtable}
\begin{subtable}{0.45\textwidth}
\centering
\begin{tabular}{llc}
\toprule
\# Rules & Order  & Acc  \\
\midrule
\multirow{3}{*}{4} & Forward & 98.5\%\\
& Backward & 95.5\%\\
& Shuffled & 94.5\%\\
\midrule
\multirow{3}{*}{5} & Forward & 97.0\%\\
& Backward & 93.5\%\\
& Shuffled & 94.8\%\\
\midrule
\multirow{3}{*}{6} & Forward & 88.0\%\\
& Backward & 85.0\%\\
& Shuffled & 88.5\%\\
\midrule
\multirow{3}{*}{7} & Forward & 87.5\%\\
& Backward & 68.0\%\\
& Shuffled & 75.8\%\\
\midrule
\multirow{3}{*}{8} & Forward & 84.5\%\\
& Backward & 63.0\%\\
& Shuffled & 66.0\%\\
\midrule
\multirow{3}{*}{9} & Forward & 81.5\%\\
& Backward & 56.5\%\\
& Shuffled & 60.8\%\\
\midrule
\multirow{3}{*}{10} & Forward & 79.5\%\\
& Backward & 46.5\%\\
& Shuffled & 55.5\%\\
\midrule
\multirow{3}{*}{11} & Forward & 73.0\%\\
& Backward & 43.5\%\\
& Shuffled & 42.5\%\\
\midrule
\multirow{3}{*}{12} & Forward & 64.0\%\\
& Backward & 32.5\%\\
& Shuffled & 38.2\%\\
\bottomrule
\end{tabular}
\caption{PaLM 2-L.}
\label{tab:logic-distractors-5-palm}
\end{subtable}
\caption{Results corresponding to Figure~\ref{fig:logic-distractors} with 5 distracting rules.}
\label{tab:logic-distractors-5}
\end{table*}
\begin{table*}[h]
\centering
\begin{subtable}{0.45\textwidth}
\centering
\begin{tabular}{llc}
\toprule
\# Rules & Order  & Acc  \\
\midrule
\multirow{3}{*}{4} & Forward & 97.0\% \\
& Backward & 98.0\% \\
& Shuffled & 97.7\% \\
\midrule
\multirow{3}{*}{5} & Forward & 98.0\% \\
& Backward & 96.0\% \\
& Shuffled & 96.5\%\\
\midrule
\multirow{3}{*}{6} & Forward & 92.5\% \\
& Backward & 88.5\%\\
& Shuffled & 90.3\%\\
\midrule
\multirow{3}{*}{7} & Forward & 84.5\% \\
& Backward & 80.0\% \\
& Shuffled & 76.0\% \\
\midrule
\multirow{3}{*}{8} & Forward & 81.5\% \\
& Backward & 76.5\% \\
& Shuffled & 70.5\% \\
\midrule
\multirow{3}{*}{9} & Forward & 73.0\% \\
& Backward & 65.0\% \\
& Shuffled & 62.8\% \\
\midrule
\multirow{3}{*}{10} & Forward & 64.5\% \\
& Backward & 59.0\% \\
& Shuffled & 53.7\% \\
\midrule
\multirow{3}{*}{11} & Forward & 58.5\% \\
& Backward & 53.0\% \\
& Shuffled & 48.7\% \\
\bottomrule
\multirow{3}{*}{12} & Forward & 57.5\% \\
& Backward & 46.5\% \\
& Shuffled & 40.0\% \\
\bottomrule
\end{tabular}
\caption{GPT-4-turbo.}
\label{tab:logic-distractors-10-gpt4}
\end{subtable}
\begin{subtable}{0.45\textwidth}
\centering
\begin{tabular}{llc}
\toprule
\# Rules & Order  & Acc  \\
\midrule
\multirow{3}{*}{4} & Forward & 97.5\%\\
& Backward & 95.0\%\\
& Shuffled & 96.3\%\\
\midrule
\multirow{3}{*}{5} & Forward & 94.0\%\\
& Backward & 91.0\%\\
& Shuffled & 92.5\%\\
\midrule
\multirow{3}{*}{6} & Forward & 89.0\%\\
& Backward & 77.0\%\\
& Shuffled & 79.7\%\\
\midrule
\multirow{3}{*}{7} & Forward & 71.5\%\\
& Backward & 55.0\%\\
& Shuffled & 60.7\%\\
\midrule
\multirow{3}{*}{8} & Forward & 68.5\%\\
& Backward & 39.5\%\\
& Shuffled & 46.7\%\\
\midrule
\multirow{3}{*}{9} & Forward & 61.5\%\\
& Backward & 38.0\%\\
& Shuffled & 42.7\%\\
\midrule
\multirow{3}{*}{10} & Forward & 47.0\%\\
& Backward & 29.5\%\\
& Shuffled & 30.7\%\\
\midrule
\multirow{3}{*}{11} & Forward & 46.5\%\\
& Backward & 15.5\%\\
& Shuffled & 25.0\%\\
\midrule
\multirow{3}{*}{12} & Forward & 36.5\%\\
& Backward & 15.5\%\\
& Shuffled & 18.2\%\\
\bottomrule
\end{tabular}
\caption{PaLM 2-L.}
\label{tab:logic-distractors-10-palm}
\end{subtable}
\caption{Results corresponding to Figure~\ref{fig:logic-distractors} with 10 distracting rules.}
\label{tab:logic-distractors-10}
\end{table*}
\begin{table*}[h]
\centering
\begin{subtable}{0.45\textwidth}
\centering
\scalebox{0.8}{
\begin{tabular}{llc}
\toprule
\# Rules & $\tau$  & Acc \\
\midrule
\multirow{5}{*}{8} & 1.0 & 99.0\%\\
& 0.5 & 95.0\%\\
& 0.0 & 91.0\%\\
& -0.5 & 94.5\%\\
& -1.0 & 95.5\%\\
\midrule
\multirow{5}{*}{10} & 1.0 & 99.0\%\\
& 0.5 & 91.0\%\\
& 0.0 & 82.5\%\\
& -0.5 & 88.5\%\\
& -1.0 & 92.5\%\\
\midrule
\multirow{5}{*}{11} & 1.0 & 98.5\%\\
& 0.5 & 90.0\%\\
& 0.0 & 84.5\%\\
& -0.5 & 88.0\%\\
& -1.0 & 91.0\%\\
\midrule
\multirow{5}{*}{12} & 1.0 & 96.5\%\\
& 0.5 & 76.0\%\\
& 0.0 & 82.0\%\\
& -0.5 & 84.5\%\\
& -1.0 & 84.0\%\\
\bottomrule
\end{tabular}}
\caption{GPT-4-turbo.}
\label{tab:logic-order-gpt4-0-distractors}
\end{subtable}
\begin{subtable}{0.45\textwidth}
\centering
\scalebox{0.8}{
\begin{tabular}{llc}
\toprule
\# Rules & $\tau$  & Acc \\
\midrule
\multirow{5}{*}{8} & 1.0 & 95.5\% \\
& 0.5 & 89.5\%\\
& 0.0 & 86.5\%\\
& -0.5 & 87.0\%\\
& -1.0 & 77.0\%\\
\midrule
\multirow{5}{*}{10} & 1.0 & 95.0\%\\
& 0.5 & 84.0\%\\
& 0.0 & 83.0\%\\
& -0.5 & 76.0\%\\
& -1.0 & 75.5\%\\
\midrule
\multirow{5}{*}{11} & 1.0 & 94.0\%\\
& 0.5 & 80.5\%\\
& 0.0 & 76.5\%\\
& -0.5 & 79.0\%\\
& -1.0 & 66.0\%\\
\midrule
\multirow{5}{*}{12} & 1.0 & 88.0\%\\
& 0.5 & 74.5\%\\
& 0.0 & 65.5\%\\
& -0.5 & 59.5\%\\
& -1.0 & 57.5\%\\
\bottomrule
\end{tabular}}
\caption{PaLM 2-L.}
\label{tab:logic-order-palm-0-distractors}
\end{subtable}
\begin{subtable}{0.45\textwidth}
\centering
\scalebox{0.8}{
\begin{tabular}{llc}
\toprule
\# Rules & $\tau$  & Acc \\
\midrule
\multirow{5}{*}{6} & 1.0 & 87.5\% \\
& 0.5 & 68.5\%\\
& 0.0 & 75.5\%\\
& -0.5 & 72.0\%\\
& -1.0 & 77.5\%\\
\midrule
\multirow{5}{*}{8} & 1.0 & 50.0\%\\
& 0.5 & 10.5\%\\
& 0.0 & 12.0\%\\
& -0.5 & 15.0\%\\
& -1.0 & 17.5\%\\
\midrule
\multirow{5}{*}{10} & 1.0 & 34.0\%\\
& 0.5 & 2.0\%\\
& 0.0 & 3.5\%\\
& -0.5 & 2.0\%\\
& -1.0 & 4.5\%\\
\midrule
\multirow{5}{*}{12} & 1.0 & 16.5\%\\
& 0.5 & 0.0\%\\
& 0.0 & 0.0\%\\
& -0.5 & 0.5\%\\
& -1.0 & 0.5\%\\
\bottomrule
\end{tabular}}
\caption{Gemini 1.0 Pro.}
\label{tab:logic-order-gemini-0-distractors}
\end{subtable}
\begin{subtable}{0.45\textwidth}
\centering
\scalebox{0.8}{
\begin{tabular}{llc}
\toprule
\# Rules & $\tau$  & Acc \\
\midrule
\multirow{5}{*}{6} & 1.0 & 87.5\% \\
& 0.5 & 68.5\% \\
& 0.0 & 75.5\%\\
& -0.5 & 72.0\% \\
& -1.0 & 77.5\%\\
\midrule
\multirow{5}{*}{8} & 1.0 & 50.0\% \\
& 0.5 & 10.5\%\\
& 0.0 & 12.0\%\\
& -0.5 & 15.0\%\\
& -1.0 & 17.5\%\\
\midrule
\multirow{5}{*}{10} & 1.0 & 34.0\%\\
& 0.5 & 2.0\%\\
& 0.0 & 3.5\%\\
& -0.5 & 2.0\%\\
& -1.0 & 4.5\%\\
\midrule
\multirow{5}{*}{12} & 1.0 & 16.5\%\\
& 0.5 & 0.0\%\\
& 0.0 & 0.0\%\\
& -0.5 & 0.5\%\\
& -1.0 & 0.5\%\\
\bottomrule
\end{tabular}}
\caption{GPT-3.5-turbo.}
\label{tab:logic-order-gpt3.5-0-distractors}
\end{subtable}
\caption{Result table corresponding to Figure~\ref{fig:logic-order-0-distractors}.}
\label{tab:logic-order-0-distractors}
\end{table*}
\begin{table*}[h]
\centering
\begin{subtable}{0.45\textwidth}
\centering
\begin{tabular}{llc}
\toprule
\# Rules & $\tau$  & Acc \\
\midrule
\multirow{5}{*}{8} & 1.0 & 89.5\%\\
& 0.5 & 86.5\%\\
& 0.0 & 78.0\%\\
& -0.5 & 82.0\%\\
& -1.0 & 85.5\%\\
\midrule
\multirow{5}{*}{10} & 1.0 & 89.0\%\\
& 0.5 & 75.5\%\\
& 0.0 & 70.5\%\\
& -0.5 & 76.5\%\\
& -1.0 & 77.0\%\\
\midrule
\multirow{5}{*}{11} & 1.0 & 84.5\%\\
& 0.5 & 68.5\%\\
& 0.0 & 67.5\%\\
& -0.5 & 78.5\%\\
& -1.0 & 75.5\%\\
\midrule
\multirow{5}{*}{12} & 1.0 & 80.5\%\\
& 0.5 & 49.5\%\\
& 0.0 & 61.5\%\\
& -0.5 & 60.5\%\\
& -1.0 & 72.5\%\\
\bottomrule
\end{tabular}
\caption{GPT-4-turbo.}
\label{tab:logic-order-5-distractors-gpt4}
\end{subtable}
\begin{subtable}{0.45\textwidth}
\centering
\begin{tabular}{llc}
\toprule
\# Rules & $\tau$  & Acc \\
\midrule
\multirow{5}{*}{8} & 1.0 & 84.5\% \\
& 0.5 & 67.5\%\\
& 0.0 & 67.0\%\\
& -0.5 & 63.5\%\\
& -1.0 & 63.0\%\\
\midrule
\multirow{5}{*}{10} & 1.0 & 79.5\%\\
& 0.5 & 58.0\%\\
& 0.0 & 56.0\%\\
& -0.5 & 52.5\%\\
& -1.0 & 46.5\%\\
\midrule
\multirow{5}{*}{11} & 1.0 & 73.0\%\\
& 0.5 & 41.5\%\\
& 0.0 & 40.0\%\\
& -0.5 & 46.0\%\\
& -1.0 & 43.5\%\\
\midrule
\multirow{5}{*}{12} & 1.0 & 64.0\%\\
& 0.5 & 39.0\%\\
& 0.0 & 42.0\%\\
& -0.5 & 33.5\%\\
& -1.0 & 32.5\%\\
\bottomrule
\end{tabular}
\caption{PaLM 2-L.}
\label{tab:logic-order-5-distractors-palm}
\end{subtable}
\caption{Results corresponding to Figure~\ref{fig:logic-order-distractors} with 5 distracting rules.}
\label{tab:logic-order-5-distractors}
\end{table*}
\begin{table*}[h]
\centering
\begin{subtable}{0.45\textwidth}
\centering
\begin{tabular}{llc}
\toprule
\# Rules & $\tau$  & Acc \\
\midrule
\multirow{5}{*}{8} & 1.0 & 81.5\% \\
& 0.5 & 73.0\%\\
& 0.0 & 65.5\%\\
& -0.5 & 73.0\% \\
& -1.0 & 76.5\% \\
\midrule
\multirow{5}{*}{10} & 1.0 & 64.5\%\\
& 0.5 & 48.5\%\\
& 0.0 & 50.5\%\\
& -0.5 & 62.0\% \\
& -1.0 & 59.0\%\\
\midrule
\multirow{5}{*}{11} & 1.0 &58.5\% \\
& 0.5 & 54.0\%\\
& 0.0 & 41.0\%\\
& -0.5 & 51.0\%\\
& -1.0 & 53.0\%\\
\midrule
\multirow{5}{*}{12} & 1.0 & 57.5\%\\
& 0.5 & 33.0\% \\
& 0.0 & 42.0\%\\
& -0.5 & 45.0\% \\
& -1.0 & 46.5\% \\
\bottomrule
\end{tabular}
\caption{GPT-4-turbo.}
\label{tab:logic-order-10-distractors-gpt4}
\end{subtable}
\begin{subtable}{0.45\textwidth}
\centering
\begin{tabular}{llc}
\toprule
\# Rules & $\tau$  & Acc \\
\midrule
\multirow{5}{*}{8} & 1.0 & 68.5\%\\
& 0.5 & 48.5\%\\
& 0.0 & 45.5\%\\
& -0.5 & 46.0\%\\
& -1.0 & 39.5\%\\
\midrule
\multirow{5}{*}{10} & 1.0 & 47.0\%\\
& 0.5 & 35.0\%\\
& 0.0 & 30.0\%\\
& -0.5 & 27.0\%\\
& -1.0 & 29.5\%\\
\midrule
\multirow{5}{*}{11} & 1.0 & 46.5\%\\
& 0.5 & 30.0\%\\
& 0.0 & 24.5\%\\
& -0.5 & 20.5\%\\
& -1.0 & 15.5\%\\
\midrule
\multirow{5}{*}{12} & 1.0 & 36.5\%\\
& 0.5 & 18.0\%\\
& 0.0 & 19.0\%\\
& -0.5 & 17.5\%\\
& -1.0 & 15.5\%\\
\bottomrule
\end{tabular}
\caption{PaLM 2-L.}
\label{tab:logic-order-10-distractors-palm}
\end{subtable}
\caption{Results corresponding to Figure~\ref{fig:logic-order-distractors} with 10 distracting rules.}
\label{tab:logic-order-10-distractors}
\end{table*}

\section{Full Results on \gsmname{}}
\label{app:exp-gsm}

Tables~\ref{tab:gsm-res-step} and~\ref{tab:gsm-res-len} present the accuracy numbers for Figures~\ref{fig:gsm-step} and~\ref{fig:gsm-len}, which are breakdown results on \gsmname{} problems with different numbers of reasoning steps and different numbers of sentences in the problem description respectively.

\begin{table*}[h]
\centering
\begin{subtable}{0.45\textwidth}
\centering
\begin{tabular}{lcc}
\toprule
\# Steps & Init Acc & Reorder Acc \\
\midrule
$>=$ 2 & 94.1\% & 85.0\% \\
$>=$ 3 & 94.0\% & 84.0\% \\
$>=$ 4 & 94.3\% & 82.8\% \\
$>=$ 5 & 92.4\% & 79.3\% \\
$>=$ 6 & 89.8\% & 73.5\% \\
\bottomrule
\end{tabular}
\caption{GPT-4-turbo.}
\label{tab:gsm-res-step-gpt4}
\end{subtable}
\begin{subtable}{0.45\textwidth}
\centering
\begin{tabular}{lcc}
\toprule
\# Steps & Init Acc & Reorder Acc \\
\midrule
$>=$ 2 & 86.4\% & 79.5\% \\
$>=$ 3 & 85.5\% & 78.5\% \\
$>=$ 4 & 84.1\% & 77.7\% \\
$>=$ 5 & 80.4\% & 71.7\% \\
$>=$ 6 & 69.4\% & 63.3\% \\
\bottomrule
\end{tabular}
\caption{PaLM 2-L.}
\label{tab:gsm-res-step-palm2}
\end{subtable}
\begin{subtable}{0.45\textwidth}
\centering
\begin{tabular}{lcc}
\toprule
\# Steps & Init Acc & Reorder Acc \\
\midrule
$>=$ 2 & 80.5\% & 69.1\% \\
$>=$ 3 & 79.0\% & 68.0\% \\
$>=$ 4 & 80.3\% & 66.2\% \\
$>=$ 5 & 80.4\% & 59.8\% \\
$>=$ 6 & 71.4\% & 55.1\% \\
\bottomrule
\end{tabular}
\caption{Gemini 1.0 Pro.}
\label{tab:gsm-res-step-gemini}
\end{subtable}
\begin{subtable}{0.45\textwidth}
\centering
\begin{tabular}{lcc}
\toprule
\# Steps & Init Acc & Reorder Acc \\
\midrule
$>=$ 2 & 67.3\% & 51.8\% \\
$>=$ 3 & 66.5\% & 51.0\% \\
$>=$ 4 & 63.1\% & 47.8\% \\
$>=$ 5 & 58.7\% & 39.1\% \\
$>=$ 6 & 42.9\% & 26.5\% \\
\bottomrule
\end{tabular}
\caption{GPT-3.5-turbo.}
\label{tab:gsm-res-step-gpt3.5}
\end{subtable}
\caption{Results corresponding to Figure~\ref{fig:gsm-step}.}
\label{tab:gsm-res-step}
\end{table*}
\begin{table*}[h]
\centering
\begin{subtable}{0.45\textwidth}
\centering
\begin{tabular}{lcc}
\toprule
\# Sentences & Init Acc & Reorder Acc \\
\midrule
$>=$ 5 & 94.1\% & 85.0\% \\
$>=$ 6 & 89.7\% & 81.6\% \\
$>=$ 7 & 86.4\% & 68.2\% \\
\bottomrule
\end{tabular}
\caption{GPT-4-turbo.}
\label{tab:gsm-res-len-gpt4}
\end{subtable}
\begin{subtable}{0.45\textwidth}
\centering
\begin{tabular}{lcc}
\toprule
\# Sentences & Init Acc & Reorder Acc \\
\midrule
$>=$ 5 & 86.4\% & 79.5\% \\
$>=$ 6 & 78.2\% & 69.0\% \\
$>=$ 7 & 77.3\% & 72.7\% \\
\bottomrule
\end{tabular}
\caption{PaLM 2-L.}
\label{tab:gsm-res-len-palm2}
\end{subtable}
\begin{subtable}{0.45\textwidth}
\centering
\begin{tabular}{lcc}
\toprule
\# Sentences & Init Acc & Reorder Acc \\
\midrule
$>=$ 5 & 80.5\% & 69.1\% \\
$>=$ 6 & 80.5\% & 60.9\% \\
$>=$ 7 & 72.7\% & 54.5\% \\
\bottomrule
\end{tabular}
\caption{Gemini 1.0 Pro.}
\label{tab:gsm-res-len-gemini}
\end{subtable}
\begin{subtable}{0.45\textwidth}
\centering
\begin{tabular}{lcc}
\toprule
\# Sentences & Init Acc & Reorder Acc \\
\midrule
$>=$ 5 & 67.3\% & 51.8\% \\
$>=$ 6 & 62.1\% & 46.0\% \\
$>=$ 7 & 54.5\% & 36.4\% \\
\bottomrule
\end{tabular}
\caption{GPT-3.5-turbo.}
\label{tab:gsm-res-len-gpt3.5}
\end{subtable}
\caption{Results corresponding to Figure~\ref{fig:gsm-len}.}
\label{tab:gsm-res-len}
\end{table*}

\end{document}